\documentclass[lettersize,journal]{IEEEtran}
\usepackage{amsmath,amsfonts}
\usepackage{algorithmic}
\usepackage{algorithm}
\usepackage{array}
\usepackage[caption=false,font=small]{subfig}
\usepackage{textcomp}
\usepackage{stfloats}
\usepackage{url}
\usepackage{verbatim}
\usepackage{graphicx}
\usepackage{cite}
\usepackage{amssymb}
\usepackage{balance}

\usepackage[switch]{lineno}
\begin{document}

\title{ODD: Omni Differential Drive for Simultaneous Reconfiguration and Omnidirectional Mobility of Wheeled Robots}

\author{Ziqi~Zhao, 
        Peijia~Xie, 
        and Max~Q.-H.~Meng,~\IEEEmembership{Fellow,~IEEE}
        % <-this % stops a space
\thanks{This work is partially supported by Shenzhen Key Laboratory of Robotics Perception and Intelligence (ZDSYS20200810171800001), Shenzhen Science and Technology Program under Grant RCBS20221008093305007, 20231115141459001, Young Elite Scientists Sponsorship Program by CAST under Grant 2023QNRC001, High level of special funds(G03034K003)from Southern University of Science and Technology, Shenzhen, China. \emph{(Corresponding author: Max~Q.-H.~Meng.)}} %<-this stops a blank
\thanks{Ziqi Zhao, Peijia Xie and Max Q.-H. Meng are with Shenzhen Key Laboratory of Robotics Perception and Intelligence, and the Department of Electronic and Electrical Engineering, Southern University of Science and Technology, Shenzhen, China.(Email: zhaozq2020@mail.sustech.edu.cn, xiepj2022@mail.sustech.edu.cn, max.meng@ieee.org)}
\thanks{Max Q.-H. Meng is also a Professor Emeritus in the Department of Electronic Engineering at The Chinese University of Hong Kong in Hong Kong and was a Professor in the Department of Electrical and Computer Engineering at the University of Alberta in Canada.}%<-this stops a blank 
\thanks{This letter has supplementary downloadable material available at
https://doi.org/xxxxxxx, provided by the authors.}
}


\maketitle

\begin{abstract}
Wheeled robots are highly efficient in human living environments. However, conventional wheeled designs, limited by degrees of freedom, struggle to meet varying footprint needs and achieve omnidirectional mobility. This paper proposes a novel robot drive model inspired by human movements, termed as the Omni Differential Drive (ODD). The ODD model innovatively utilizes a lateral differential drive to adjust wheel spacing without adding additional actuators to the existing omnidirectional drive. This approach enables wheeled robots to achieve both simultaneous reconfiguration and omnidirectional mobility. Additionally, a prototype was developed to validate the ODD, followed by kinematic analysis. Control systems for self-balancing and motion were designed and implemented. Experimental validations confirmed the feasibility of the ODD mechanism and the effectiveness of the control strategies. The results underline the potential of this innovative drive system to enhance the mobility and adaptability of robotic platforms.
\end{abstract}


\section{Introduction}
\IEEEPARstart{T}{he} rapid advancement of robotics technology has led to its widespread application in many areas of human life. Wheeled robots have distinct advantages and disadvantages compared to legged robots. Wheeled robots are generally more efficient and faster on smooth surfaces, while legged robots excel on complex terrains.\cite{raj_comprehensive_2022,rubio_review_2019,bigdog_2023}. However, most environments where mobile robots are used in human life involve smooth surfaces. For instance, guidance robots, disinfection robots, cleaning robots, and delivery robots primarily operate in hotels, restaurants, airports, office buildings, and residential settings. These applications underscore the importance of optimizing wheeled robot designs for such environments.

Enhancing wheeled robot performance requires balancing three critical properties: passability, agility, and stability. Passability is the ability to navigate narrow spaces, requiring a small footprint. agility refers to omnidirectional movement, allowing the robot to maneuver freely in any direction. Stability requires a larger footprint to maintain balance and prevent tipping. However, these three properties often conflict. Designing a wheeled robot with passability, agility, and stability is a significant challenge.

Static stability is typically achieved through the support polygon, defined by the wheels' contact points with the ground. The size and shape of the support polygon determine static stability\cite{vukobratovic_zero-moment_2004}. However, increasing the footprint to enlarge the support polygon can negatively affect the robot's passability in narrow spaces. This issue is often addressed through reconfiguration, allowing dynamic footprint adjustment. A support polygon requires at least three non-collinear points. 

Wheel-legged robots adjust the support polygon using leg joints\cite{yun_development_2021,li_DesignControlTransformable_2024,hou_2023,fuchs_rollin_2009}. Yun et al. proposed a wheel-legged robot that alters the support polygon using leg joints while maintaining the orientation of the four Mecanum wheels via a parallel link mechanism in the legs\cite{yun_development_2021}. Li et al. introduced a multi-mode robot that adjusts the support polygon size via leg joints and switches to collinear Mecanum wheels configurations to reduce the footprint\cite{li_DesignControlTransformable_2024}. Some studies add actuators specifically to adjust the wheel spacing and change the support polygon size\cite{karamipour_reconfigurable_2020,hayat_Panthera_2019,rayguru_Introducing_2024}. Karamipour et al. proposed a mobile robot with four omni-wheels, adding a linear actuator between the wheels on each side to adjust the spacing\cite{karamipour_reconfigurable_2020}. 

Another reconfiguration method uses only wheels drive, with the body having passive degrees of freedom\cite{karamipour_omnidirectional_2019, pankert_design_2022,labazanova_SoftRigid_2023}. Karamipour et al. proposed a mobile robot with four Mecanum wheels and transverse prismatic joints between the wheels on each side to adjust the form through forces generated by the Mecanum wheels' rotation\cite{karamipour_omnidirectional_2019}. Inspired by the concept that reconfiguration uses only wheel drive, the ODD model for omnidirectional wheeled mobility was proposed, which achieves reconfiguration without the need for additional drives.

Reducing support points or using collinear arrangements results in a smaller footprint and better passability, while requiring self-balancing to maintain dynamic stability. Ball-wheel robots achieve omnidirectional movement with single-point support \cite{lauwers_dynamically_2006,chen_design_2013,minniti_whole-body_2019,kumagai_development_2008}. Single omnidirectional wheels add driven rollers to conventional wheels, enabling lateral movement \cite{shen_omburo_2020}. Systems with single-point support tend to oscillate under external disturbances and cannot remain stationary, requiring high driving torque and energy consumption. In contrast, two-point or multiple collinear support configurations achieve dynamic stability in one direction through self-balancing and static stability in another. Examples include the dual-ball mechanism \cite{gao_dual_ball_2022} and the collinear Mecanum wheel approach \cite{Reynolds-Haertle2011,miyakoshi_omnidirectional_2017,watson_collinear_2021}. In this form, the strength of static stability depends on the distance between the support points. Therefore, the objective of our work is to reconfigure that distance.

\begin{figure}[!t]
    \centering
    \includegraphics[width=3.1in]{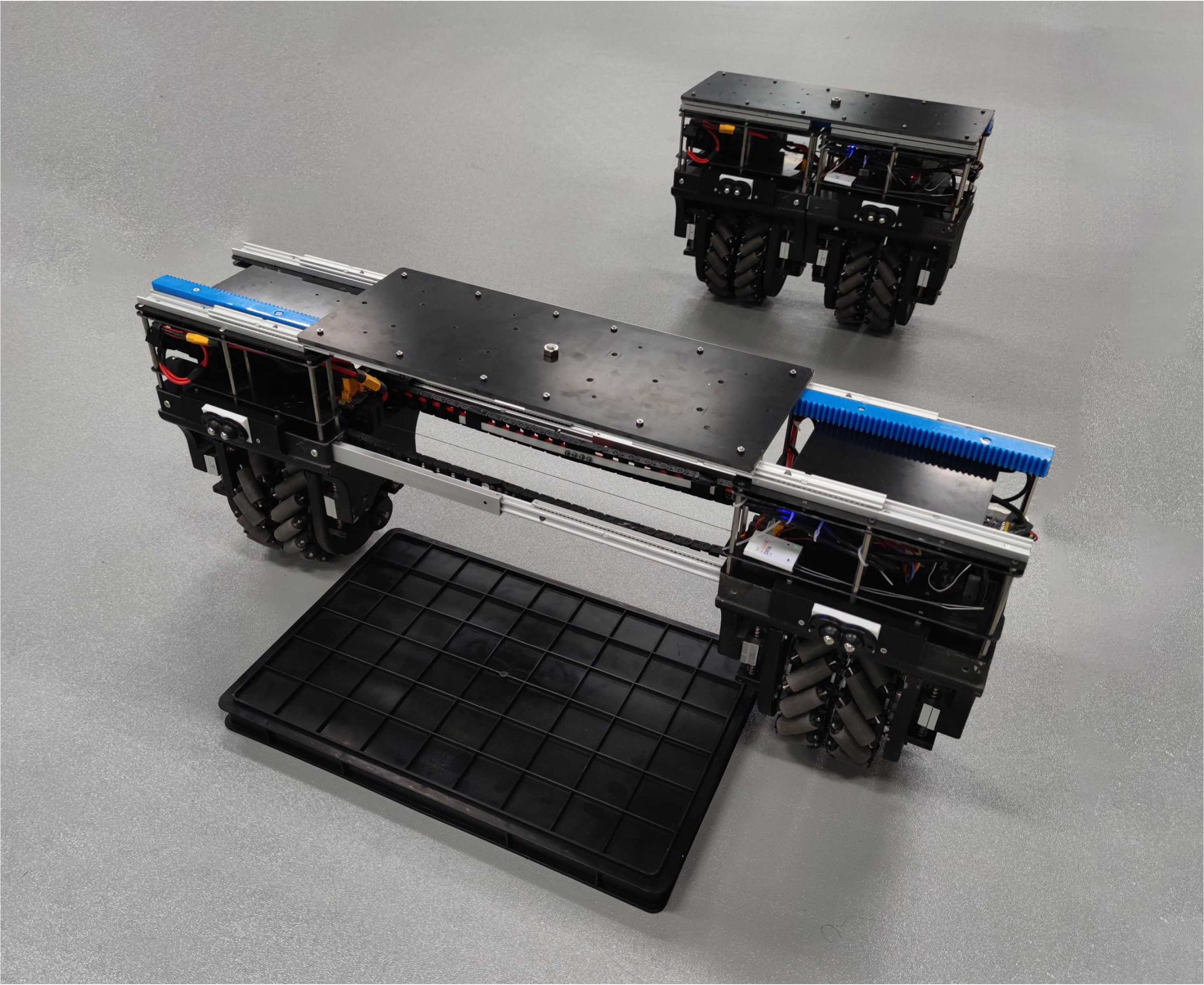}
    \caption{Proposed Prototype which can simultaneous reconfigure and omnidirectional mobile using the Omni Differential Drive (ODD).}
    \label{fig_home}
\end{figure}

In this letter, we propose the ODD, inspired by human and biped robot locomotion, to achieve passability, agility, and stability in wheeled robots. This mobility method enables omnidirectional movement and reconfiguration of wheeled mobile platforms. To validate this mobility method, we designed a prototype, as shown in Fig. \ref{fig_home}. The main contributions of this study are summarized as follows:

\begin{enumerate}
\item {An ODD wheeled mobile model was proposed, allowing the robot to achieve simultaneous omnidirectional movement and reconfiguration using only its wheels.}
\item {A prototype based on a collinear Mecanum wheel mechanism was developed to implement the ODD model, with its kinematics modeled.}
\item {Controllers were designed and implemented to operate the prototype, validating the models of both the ODD and the prototype.}
\end{enumerate}

The remainder of this paper is organized as follows: Section II introduces the concept. Section III presents the Omni Differential Drive. Section IV describes the prototype design. Section V discusses the modeling and control of the prototype. Section VI details the experimentation and validation. Finally, Section VII concludes the paper and suggests directions for future research.

\section{Concept}

\subsection{Inspired by Human Movements}

Humans, with the extensive freedom of their legs, can achieve omnidirectional movement and flexibly change direction and orientation.	 They can turn sideways to navigate narrow passages in kitchens, restaurants, or crowded areas. By widening their stance, humans can enlarge the support polygon and lower the center of gravity, enhancing stability especially when dealing with external forces during activities like boxing and Kung Fu, or when experiencing tilting and shaking in transportation modes like airplanes, trains and ships. They can step over small obstacles, like scattered items or puddles. Thus, humans possess excellent passability, stability, and agility.

\begin{figure}[!t]
    \centering
    \subfloat[]{\includegraphics[height=1.4in]{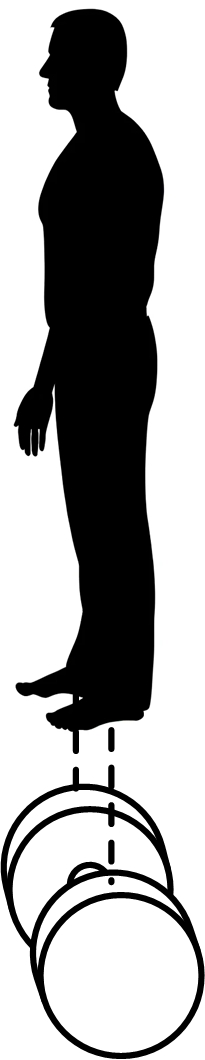}%
    \label{fig_concept_side}}
    \hspace{2pt}
    \subfloat[]{\includegraphics[height=1.4in]{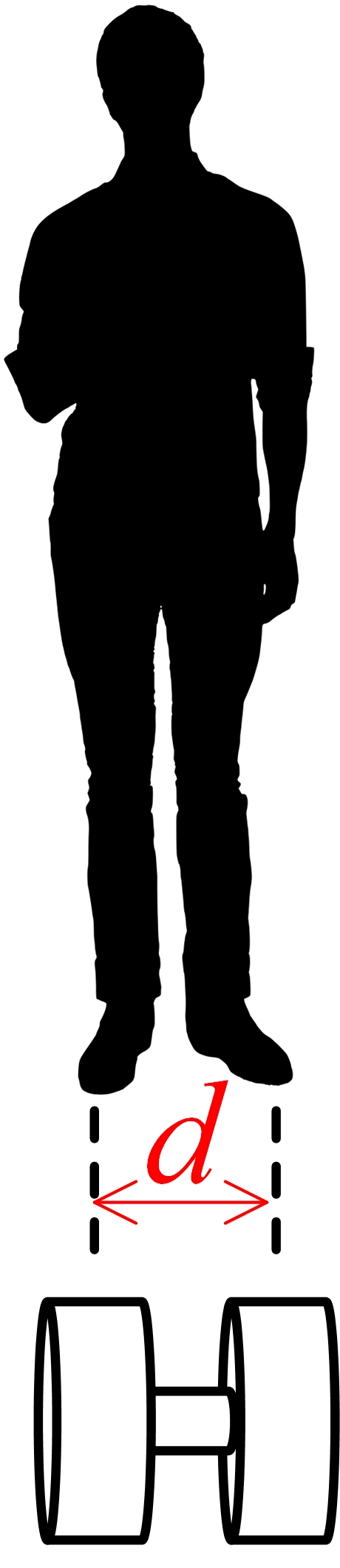}%
    \label{fig_concept_front}}
    \hspace{2pt}
    \subfloat[]{\includegraphics[height=1.4in]{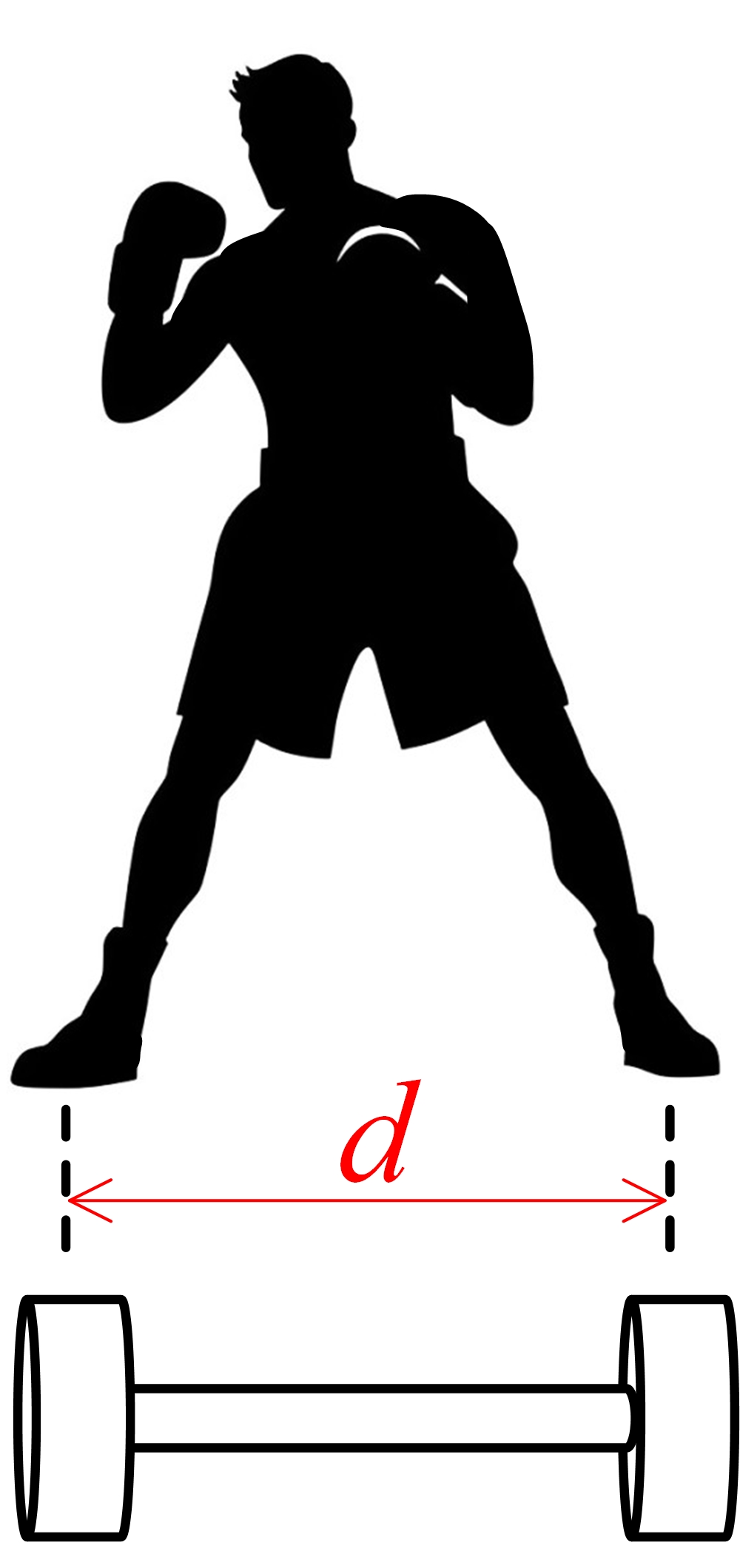}%
    \label{fig_concept_boxing}}
    \hspace{0pt}
    \subfloat[]{\includegraphics[height=1.4in]{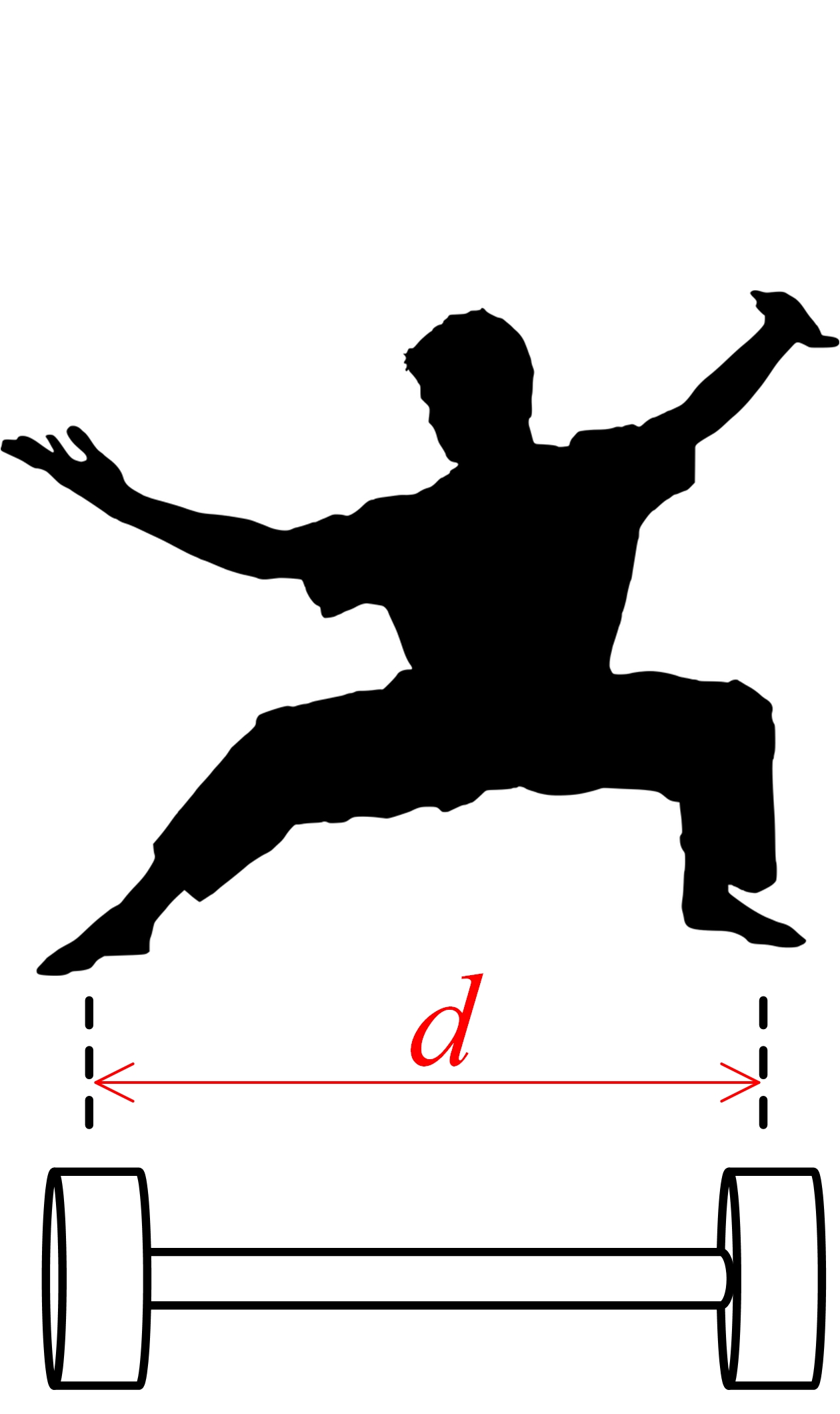}%
    \label{fig_concept_kungfu}}
    \hspace{2pt}
    \subfloat[]{\includegraphics[height=1.4in]{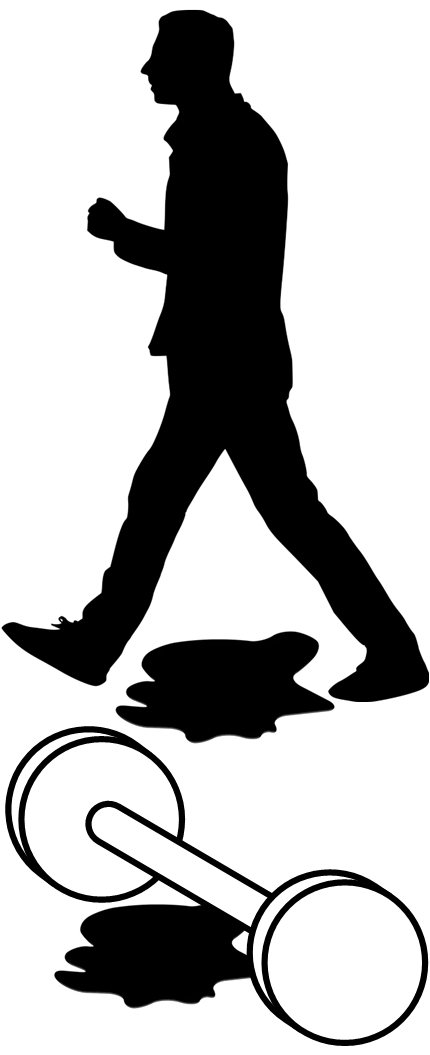}%
    \label{fig_obstacle}}
    \caption{Analogy between human movements and wheeled mobility. (a) Side-view standing or lateral walking. (b) Front-view Standing or longitudinal walking. (c) Boxing. (d) Kung Fu. (e) Obstacle crossing.}
    \label{fig_concept}
\end{figure}

Analogously, the distance between a human's feet is similar to the wheel spacing of a wheeled platform, as shown in Fig. \ref{fig_concept}. For passability, a smaller wheel spacing results in a smaller footprint, while a larger spacing helps overcome obstacles. Omnidirectional movement allows sideways navigation through narrow passages. For stability, a larger wheel spacing provides greater anti-overturning torque, maintaining static stability. Omnidirectional movement also enables dynamic stability in all directions. For agility, omnidirectional movement allows rapid direction and position changes. Thus, to meet all three properties, the mobile platform must be reconfigurable in wheel spacing and capable of omnidirectional movement.

\subsection{Robotics Applications}

The ODD method enables both omnidirectional movement and reconfiguration, making it suitable for various mobile platforms. For example, replacing the driven wheels in a two-wheel drive platform with two caster wheels and removing the wheel spacing constraint transforms it into a four-wheel mobile platform capable of changing its body width. Similarly, replacing drive wheels on a two-wheeled self-balancing vehicle with omnidirectional wheels and removing the wheel spacing constraint enables omnidirectional movement and reconfiguration. This drive method can also be applied to wheel-legged robots, scooters, and roller skates.

\section{Omni Differential Drive}
\subsection{Kinematics Model}

\begin{figure}[!t]
    \centering
    \subfloat[]{\includegraphics[height=1.15in]{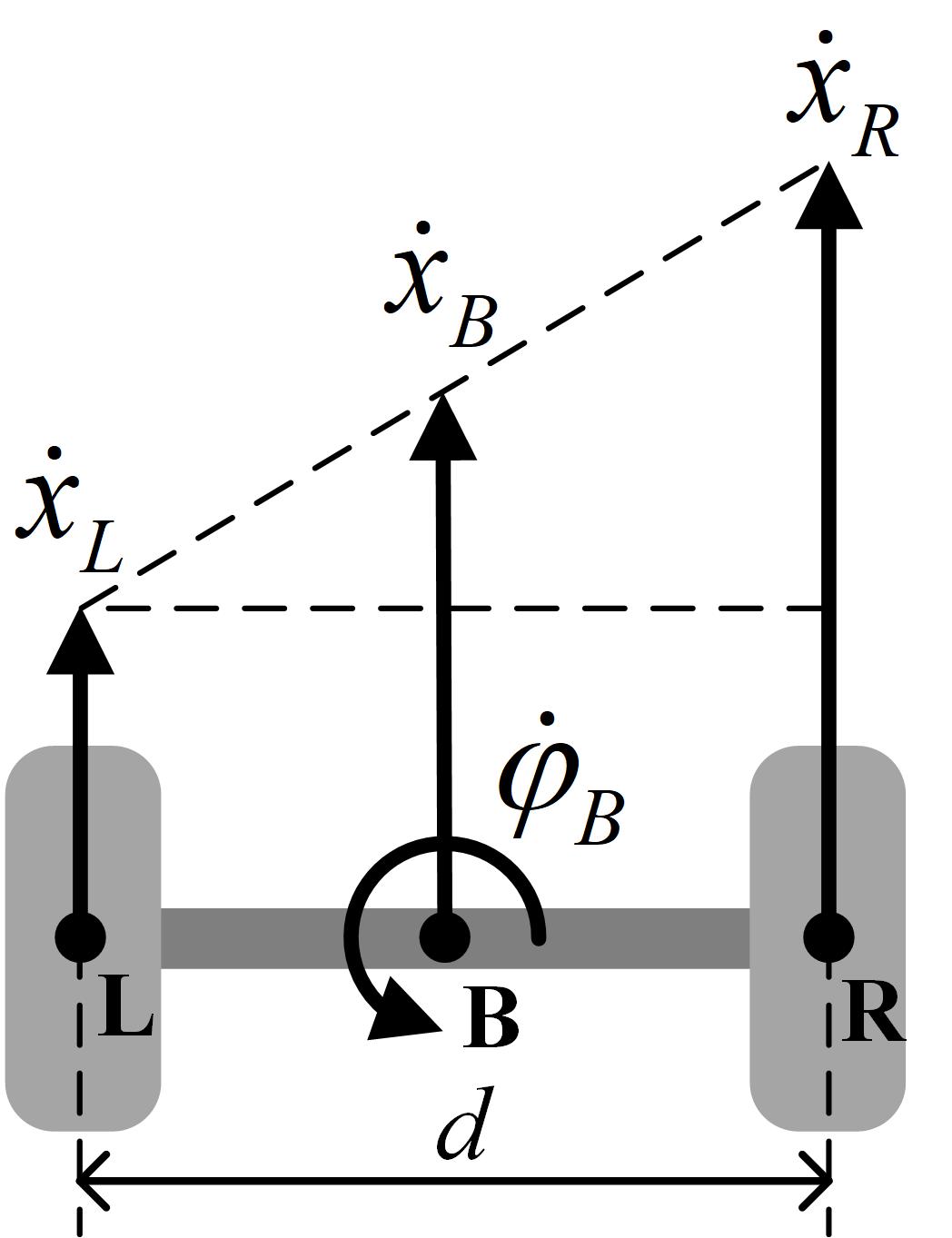}%
    \label{fig_DD}}
    \hspace{0pt}
    \subfloat[]{\includegraphics[height=1.15in]{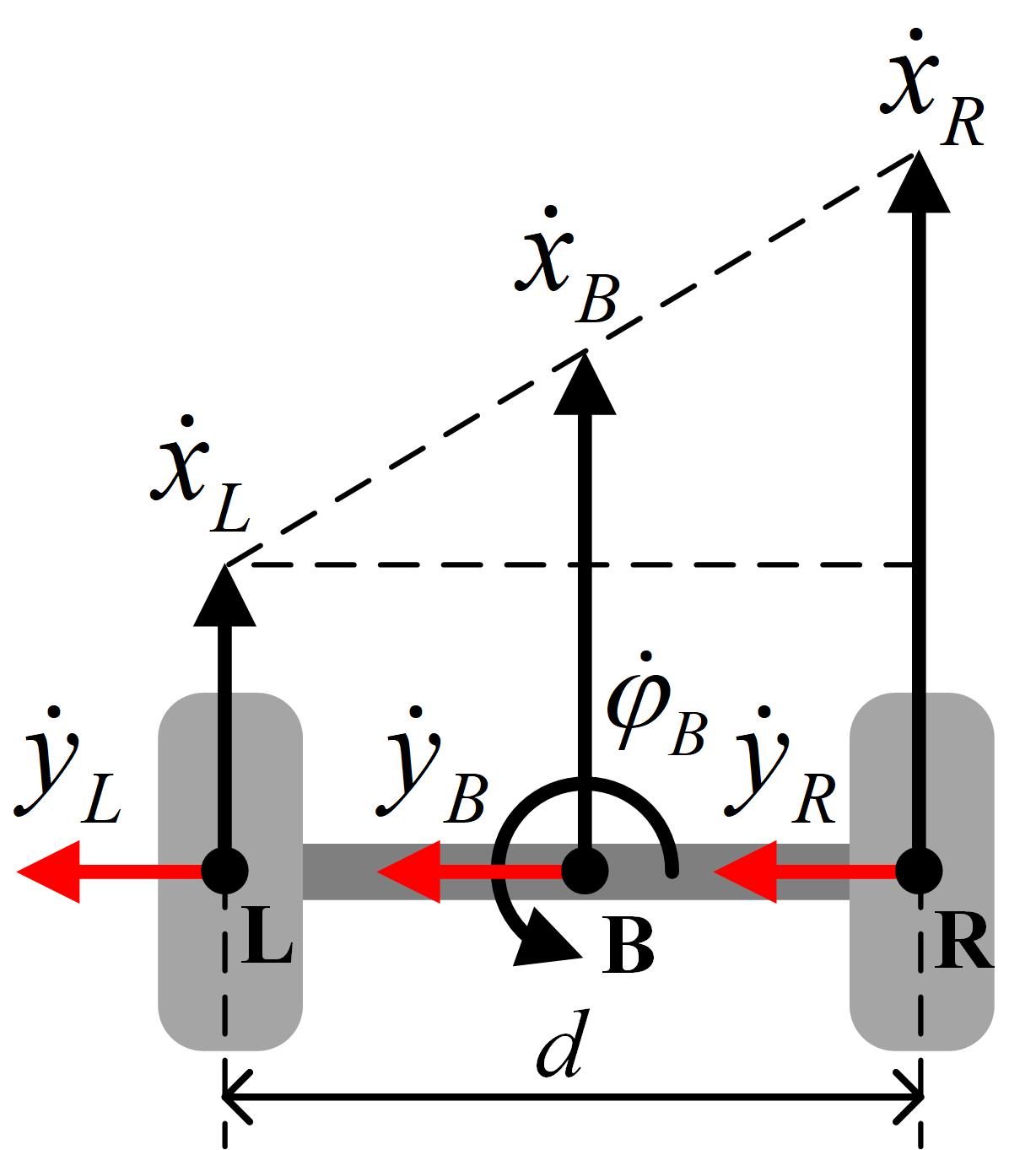}%
    \label{fig_OD}}
    \hspace{0pt}
    \subfloat[]{\includegraphics[height=1.15in]{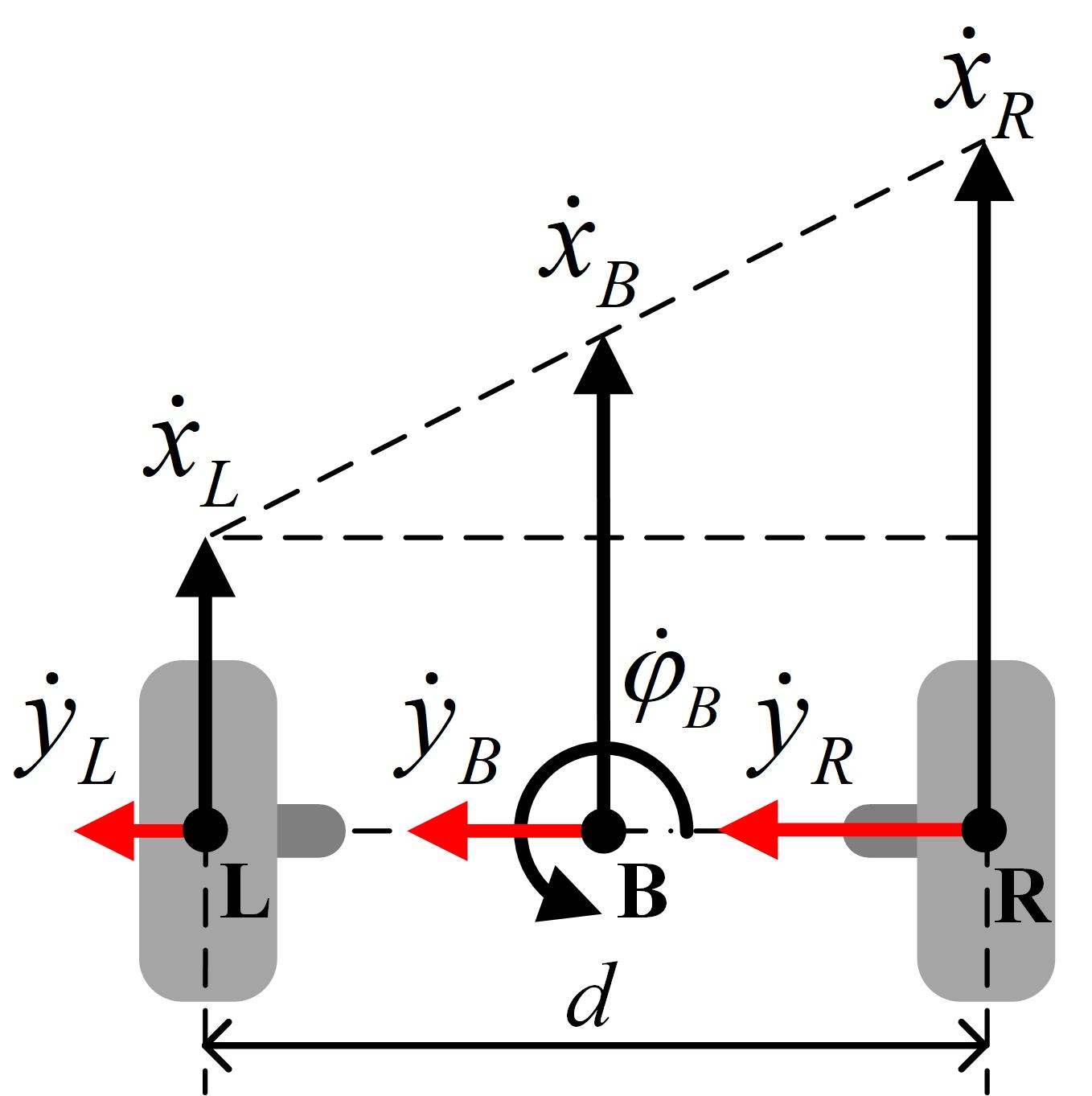}%
    \label{fig_ODD}}
    \caption{Models of drive methods. (a) Differential Drive (DD). (b) Omnidirectional Drive (OD). (c) Proposed Omni Differential Drive (ODD).}
    \label{fig_dd}
\end{figure}
 
Traditional differential drive consists of two single-degree-of-freedom wheels, as shown in Fig. \ref{fig_dd}\subref{fig_DD}. By controlling the linear velocities of the left and right wheels, $\dot{x}_L$ and $\dot{x}_R$, the linear velocity $\dot{x}_B$ and angular velocity $\dot{\varphi}_B$ of the center of the mobile platform can be controlled. The forward and inverse kinematics of the traditional differential drive are:

\begin{equation}
    \label{eqn_DD_1}
    \begin{bmatrix}
    \dot{x}_B \\
    \dot{\varphi}_B
    \end{bmatrix}=\begin{bmatrix}
    1/2 & 1/2 \\
    -1/d & 1/d
    \end{bmatrix}\begin{bmatrix}
    \dot{x}_L \\
    \dot{x}_R
    \end{bmatrix},
\end{equation}

\begin{equation}
    \label{eqn_DD_2}
    \begin{bmatrix}
    \dot{x}_L \\
    \dot{x}_R
    \end{bmatrix}=\begin{bmatrix}
    1 & -d/2 \\
    1 & d/2
    \end{bmatrix}\begin{bmatrix}
    \dot{x}_B \\
    \dot{\varphi}_B
    \end{bmatrix}.
\end{equation}

The traditional differential drive model is limited in lateral movement because conventional wheels have a single rolling degree of freedom. Replacing the differential drive wheels with omnidirectional wheels, which allow both longitudinal and lateral movement, as shown in Fig. \ref{fig_dd}\subref{fig_OD}, enables control of the linear velocities $\dot{x}_B$ and $\dot{y}_B$, and the angular velocity $\dot{\varphi}_B$ of the center of the mobile platform through $\dot{x}_L$, $\dot{y}_L$, $\dot{x}_R$, and $\dot{y}_R$. This allows the robot to achieve omnidirectional mobility. The mechanisms developed by some researchers fit this model\cite{watson_collinear_2021, gao_dual_ball_2022}. The forward and inverse kinematics of the omnidirectional drive model are:

\begin{equation}
    \label{eqn_OD_1}
    \begin{bmatrix}
    \dot{x}_B \\
    \dot{y}_B \\
    \dot{\varphi}_B
    \end{bmatrix}=\begin{bmatrix}
    1/2 & 0 & 1/2 & 0 \\
    0 & 1/2 & 0 & 1/2 \\
    -1/d & 0 & 1/d & 0
    \end{bmatrix}\begin{bmatrix}
    \dot{x}_L \\
    \dot{y}_L \\
    \dot{x}_R \\
    \dot{y}_R
    \end{bmatrix},
\end{equation}

\begin{equation}
    \label{eqn_OD_2}
    \begin{bmatrix}
    \dot{x}_L \\
    \dot{y}_L \\
    \dot{x}_R \\
    \dot{y}_R
    \end{bmatrix}=\begin{bmatrix}
    1 & 0 & -d/2 \\
    0 & 1 & 0 \\
    1 & 0 & d/2\\
    0 & 1 & 0
    \end{bmatrix}\begin{bmatrix}
    \dot{x}_B \\
    \dot{y}_B \\
    \dot{\varphi}_B
    \end{bmatrix}.
\end{equation}

As shown in (\ref{eqn_OD_1}) and (\ref{eqn_OD_2}), this system is over-actuated, requiring $\dot{y}_L$ and $\dot{y}_R$ to be equal. With a fixed wheel spacing $d$, any difference between $\dot{y}_L$ and $\dot{y}_R$ causes slipping or loss of control. Therefore, this paper proposes ODD based on OD, allowing the wheel spacing $d$ to vary. By using the difference between $\dot{y}_L$ and $\dot{y}_R$, the change in $d$ can be controlled, as shown in Fig. \ref{fig_dd}\subref{fig_ODD}. This drive model allows for the control of $\dot{x}_B$, $\dot{y}_B$, and $\dot{\varphi}_B$ of the platform's center, as well as the rate of change $\dot{d}$ of $d$ through $\dot{x}_L$, $\dot{y}_L$, $\dot{x}_R$, and $\dot{y}_R$. The forward and inverse kinematics are:

\begin{equation}
    \label{eqn_ODD_1}
    \begin{bmatrix}
    \dot{x}_B \\
    \dot{y}_B \\
    \dot{\varphi}_B \\
    \dot{d}
    \end{bmatrix}=\begin{bmatrix}
    1/2 & 0 & 1/2 & 0 \\ 
    0 & 1/2 & 0 & 1/2 \\   
    -1/d & 0 & 1/d & 0 \\
    0 & 1 & 0 & -1
    \end{bmatrix}\begin{bmatrix}
    \dot{x}_L \\
    \dot{y}_L \\
    \dot{x}_R \\
    \dot{y}_R
    \end{bmatrix},
\end{equation}

\begin{equation}
    \label{eqn_ODD_2}
    \begin{bmatrix}
    \dot{x}_L \\
    \dot{y}_L \\
    \dot{x}_R \\
    \dot{y}_R
    \end{bmatrix}=\begin{bmatrix}
    1 & 0 & -d/2 & 0 \\
    0 & 1 & 0 & 1/2 \\
    1 & 0 & d/2 & 0\\
    0 & 1 & 0 & -1/2
    \end{bmatrix}\begin{bmatrix}
    \dot{x}_B \\
    \dot{y}_B \\
    \dot{\varphi}_B \\
    \dot{d}
    \end{bmatrix}.
\end{equation}

The ODD is fully actuated and, compared to the OD, requires no additional actuators to achieve both omnidirectional movement and reconfiguration.

\subsection{Dynamics Model}

\begin{figure}[t]
    \centering
    \includegraphics[width=2.2in]{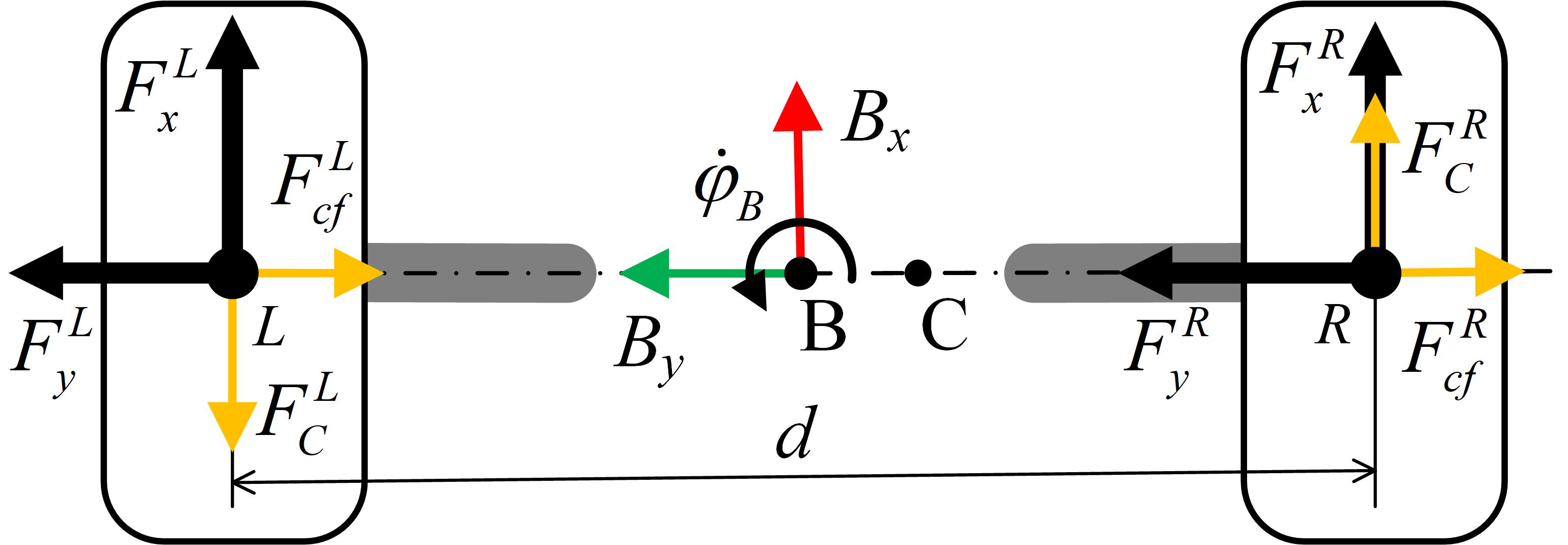}
    \caption{Force diagram of ODD.}
    \label{fig_dynamics}
\end{figure}

A dynamic model is established to investigate the performance of the ODD, and the force diagram is shown in Fig. \ref{fig_dynamics}. The left and right wheel masses are $m_L$ and $m_R$, with positions $y_L$ and $y_R$ on the $By$ axis. The center of mass is at point $C$, and its position $y_C$ is calculated as in (\ref{eqn_yc}). Subsequently, the moment of inertia $I$ can be determined as shown in (\ref{eqn_I}), where $m = m_L + m_R$. The moment of inertia $I$ is proportional to the square of $d$.

\begin{equation}
\label{eqn_yc}
y_C=\frac{m_L y_L + m_R y_R}{m_L+m_R}=\frac{(m_L-m_R)d}{2m}.
\end{equation}

\begin{equation}
\label{eqn_I}
I = m_L(y_C-y_L)^2 + m_R(y_C-y_R)^2=\frac{m_L m_R d^2}{m}.
\end{equation}

The ODD model provides driving forces $F_x$ and $F_y$ in the $x_B$ and $y_B$ directions, respectively, as well as the generated Coriolis force $F_C$, and centrifugal force $F_{cf}$. The forces on the left and right wheels are denoted by the superscripts $L$ and $R$, respectively, as shown in (\ref{eqn_dynamics0}). By calculating the resultant forces in different directions, the forward dynamics equation (\ref{eqn_dynamics1}) is obtained. Solving it leads to the inverse dynamics equation (\ref{eqn_dynamics2}).

\begin{equation}
\label{eqn_dynamics0}
\begin{aligned}
&F_C^L = m_L \dot{\varphi}_B (2\dot{y}_B-\dot{d}), 
&&F_{cf}^L = m_L \dot{\varphi}_B^2(\dot{x}_B/\dot{\varphi}_B-d/2), \\
&F_C^R = m_R \dot{\varphi}_B (2\dot{y}_B+\dot{d}),
&&F_{cf}^R = m_R \dot{\varphi}_B^2(\dot{x}_B/\dot{\varphi}_B+d/2).
\end{aligned}
\end{equation}

\begin{equation}
\label{eqn_dynamics1}
\begin{aligned}
&m\Ddot{x}_B = F_x^L-F_C^L+F_x^R+F_C^R,\\
&m\Ddot{y}_B = F_y^L-F_{cf}^L+F_y^R-F_{cf}^R,\\
&I\Ddot{\varphi}_B = (F_x^R+F_C^R)m_L d /m-(F_x^L-F_C^L)m_R d /m,\\
&\Ddot{d} = (F_y^R-F_{cf}^R)/m_R - (F_y^L-F_{cf}^L)/m_L.
\end{aligned}
\end{equation}

\begin{equation}
\label{eqn_dynamics2}
\begin{aligned}
  &F_x^L = m_L\Ddot{x}_B - I\Ddot{\varphi}_B/d + m_L (2\dot{y}_B-\dot{d})\dot{\varphi}_B,\\
&F_y^L = m_L \Ddot{y}_B - m_L m_R\Ddot{d}/m + m_L \dot{\varphi}_B^2(\dot{x}_B/\dot{\varphi}_B-d/2),\\
&F_x^R = m_R\Ddot{x}_B + I\Ddot{\varphi}_B/d - m_R (2\dot{y}_B+\dot{d})\dot{\varphi}_B,\\
&F_y^R = m_R \Ddot{y}_B - m_L m_R\Ddot{d}/m + m_R \dot{\varphi}_B^2(\dot{x}_B/\dot{\varphi}_B+d/2).
\end{aligned}
\end{equation}

\section{Prototype Design}
A prototype was designed to validate the effectiveness and accuracy of the proposed ODD, as shown in Fig. \ref{fig_prototype}. The main components include:

\begin{figure}[t]
    \centering
    \subfloat[]{\includegraphics[width=3.2in]{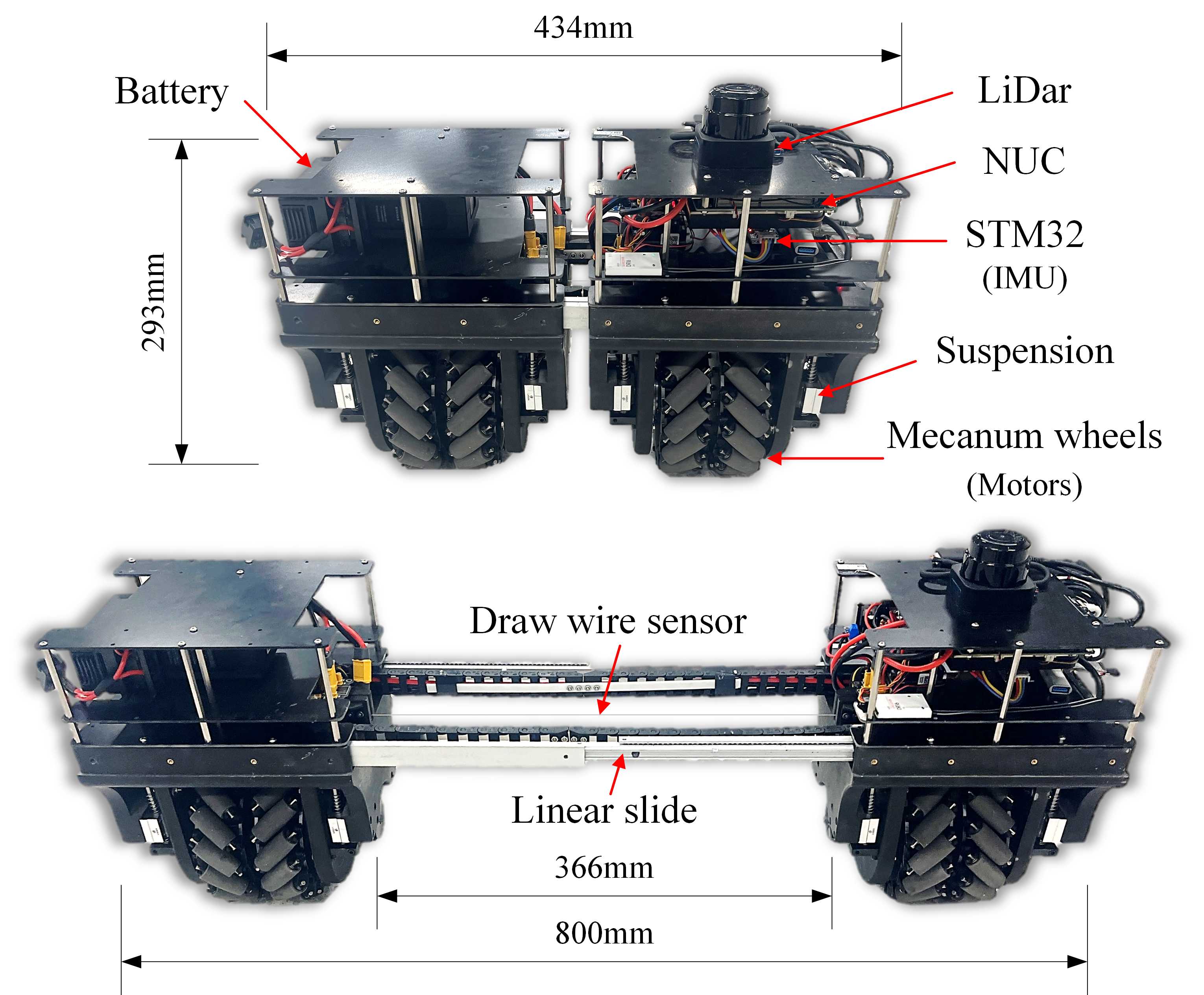}%
    \label{fig_structure}}
    \hspace{0pt}
    \subfloat[]{\includegraphics[width=3.2in]{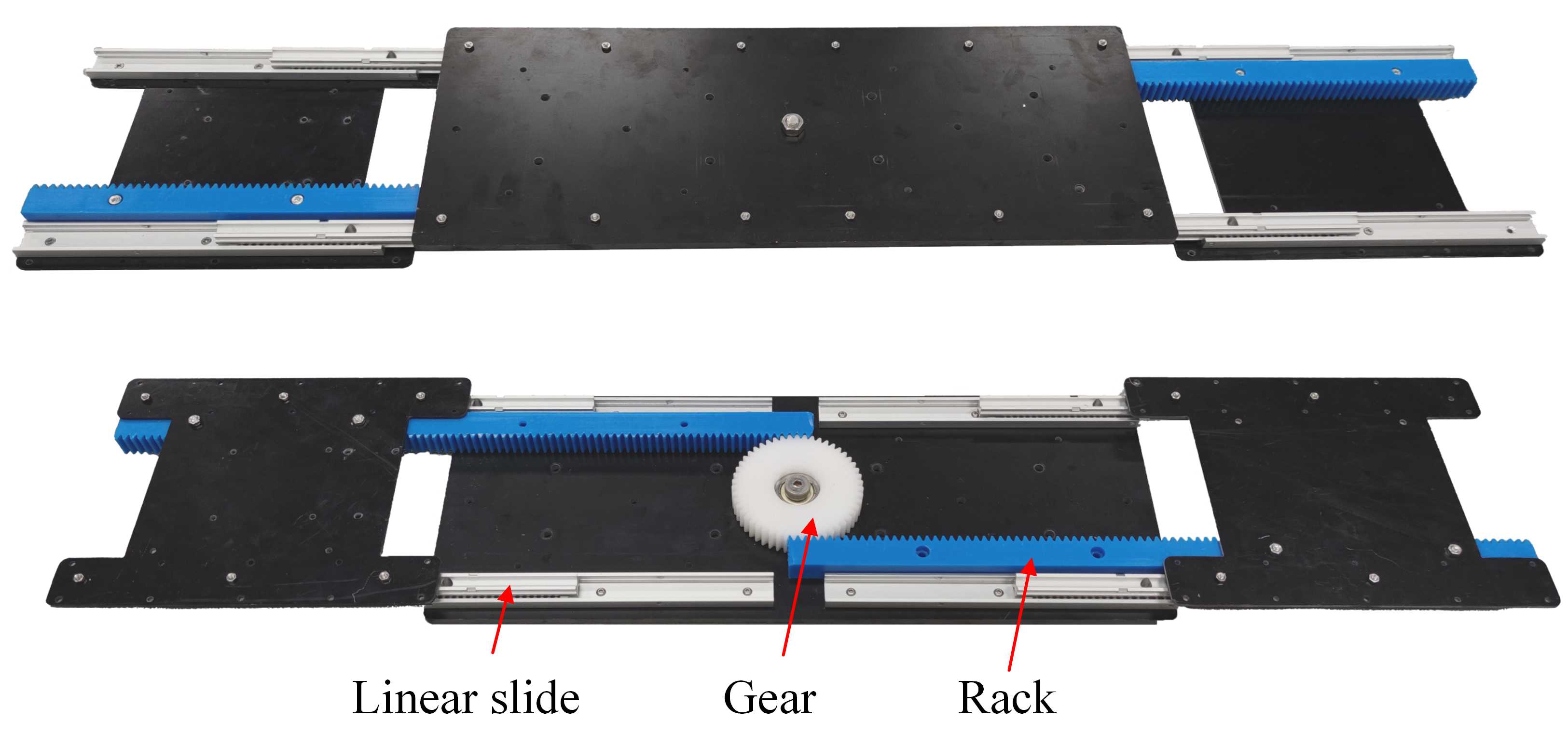}%
    \label{fig_centering}}
    \caption{Components of proposed prototype. (a) Overall structure and dimensions. (b) Self-centering platform.}
    \label{fig_prototype}
\end{figure}

\subsubsection{Active Omnidirectional Wheel}
The active omnidirectional wheel is essential for the ODD. Several solutions exist, including collinear Mecanum wheels\cite{watson_collinear_2021}, single-layer omnidirectional wheels\cite{shen_omburo_2020}, and ball wheels\cite{chen_design_2013,gao_dual_ball_2022}. This prototype uses collinear Mecanum wheels for simplicity and strong driving force. Each Mecanum wheel is powered by a DJI M3508 DC motor.
\subsubsection{Suspension System}
Each wheel has an independent suspension system to ensure ground contact on uneven surfaces.
\subsubsection{Linear Slides}
The four Mecanum wheels are divided into left and right groups, connected by two passive linear slides to maintain collinear while allowing variable wheel spacing.
\subsubsection{Self-Centering Platform}
Sensors, batteries, computers, and other loads can be placed on the mounting platforms above the left and right wheel groups, as shown in Fig. \ref{fig_prototype}\subref{fig_structure}. Additionally, a self-centering platform can be installed, which remains centered using a rack-and-pinion mechanism, as shown in Fig. \ref{fig_prototype}\subref{fig_centering}.
\subsubsection{Sensors}
A draw-wire sensor measures wheel spacing with ±0.1\% accuracy. An Inertial Measurement Unit (IMU) integrated into the STM32 microcontroller measures angles, angular velocities, and accelerations. Moreover, The motor encoders detect the motor's rotation angle and speed, while the motor current can also be monitored. Additional sensors like Light Detection and Ranging (LiDAR) and cameras, can be added to the reserved mounting platform as needed.
\subsubsection{Other Components}
The proposed prototype uses an Intel NUC computer as the upper-level controller and an STM32 as the lower-level controller, along with the motor driver boards. Additionally, a 5700mAh battery is included to ensure the robot has sufficient endurance.

\section{Modeling and Control}
\subsection{Kinematics Model}

\begin{figure}[t]
\centering
\includegraphics[width=3in]{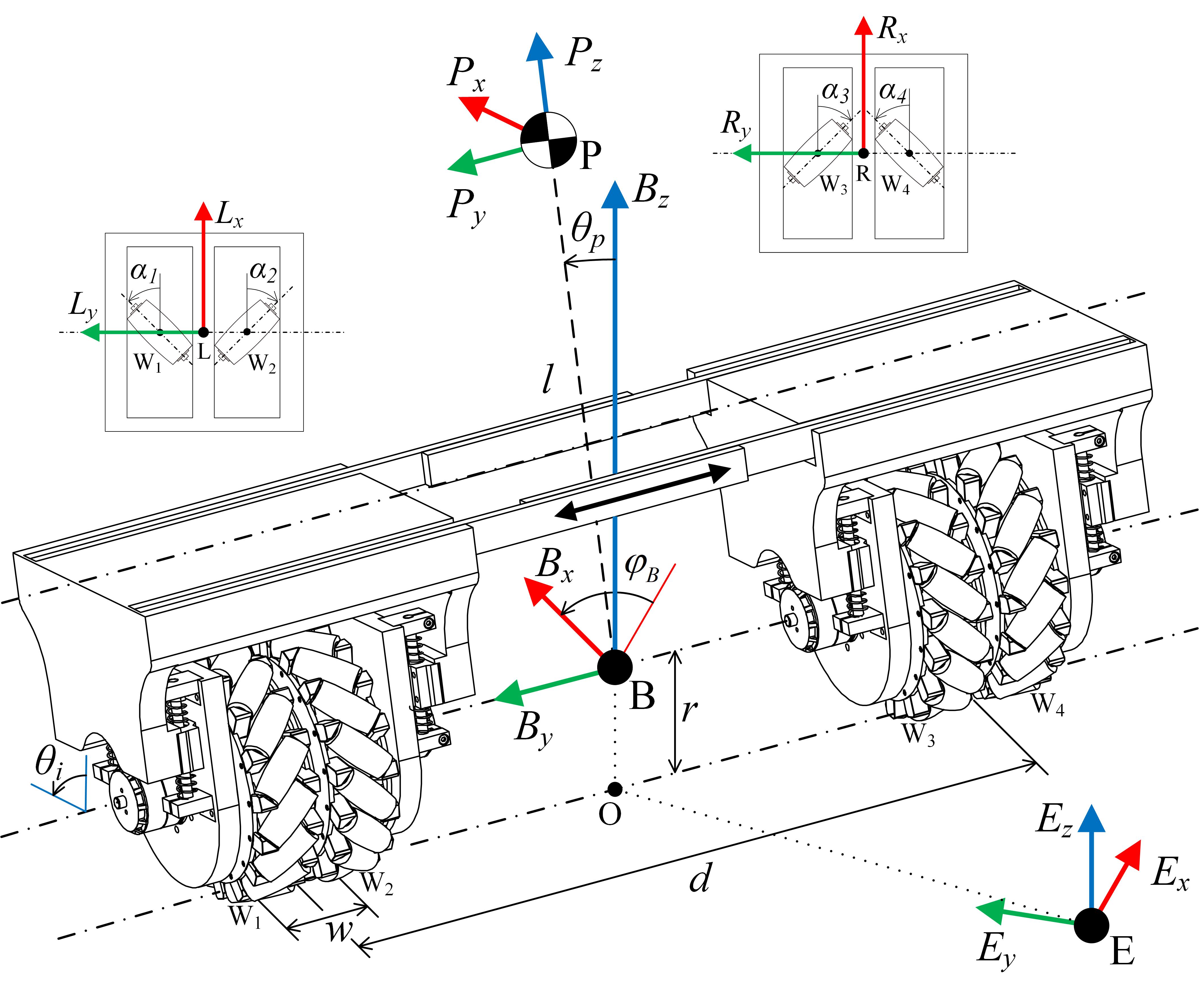}
\caption{Coordinates and parameters for the proposed prototype.}
\label{fig_kinematics}
\end{figure}

First, we model the proposed prototype using its coordinates and parameters, as shown in Fig. \ref{fig_kinematics}. The global coordinate system is denoted as $E$, and the moving platform's body coordinate system is denoted as $B$. The origin of coordinate system $B$ is located on the wheel axis, and $B$ is rotated by an angle $\varphi_B$ around the $z$-axis relative to $E$, i.e., the angle between $E_x$ and $B_x$. The projection of point $B$ onto the $E_{xy}$ plane is point $O$. The four wheels are numbered $i$ from left to right as 1, 2, 3, and 4. The contact points of the wheels with the $E_{xy}$ plane, i.e., the ground, are denoted as $W_i$. The midpoint of $W_1$ and $W_2$ is $L$, and the midpoint of $W_3$ and $W_4$ is $R$. The distance between $L$ and $R$ is $d$. The distance between $W_1$ and $W_2$, as well as between $W_3$ and $W_4$, is $w$. The radius of all four wheels is $r$, and their angular positions are $\theta_i$. The angle between the roller axis of each Mecanum wheel and the $B_x$ direction is $\alpha_i$.

Typically, modeling Mecanum wheel structures involves directly establishing the kinematic relationships between the four Mecanum wheels and the central point\cite{zimmermann_mecanum_2014,watson_collinear_2021,yun_development_2021}. However, since omnidirectional movement can be achieved through various means, many studies have realized it using different approaches\cite{gao_dual_ball_2022}. In this paper, to demonstrate the universality of the proposed ODD model, we model the prototype using the ODD model. Based on the aforementioned typical modeling approach for Mecanum wheel vehicles, establishing the motion relationships between points L and R and the four wheels, while considering $\dot{d}$. yielding equations (\ref{eqn_kinematic_L}), and then computing their inverses to obtain the inverse kinematics equation (\ref{eqn_kinematic_LR}). Substituting equation (\ref{eqn_kinematic_LR}) into the ODD motion equation (\ref{eqn_ODD_1}) yields the kinematic equation (\ref{eqn_kinematic}).In these equations, $S_i$, $C_i$, and $T_i$ represent $\sin{\alpha_i}$, $\cos{\alpha_i}$, and $\tan{\alpha_i}$, respectively.

\begin{figure*}[t]
\begin{equation}
    \label{eqn_kinematic_L}
    \begin{bmatrix}
    r\dot{\theta}_1 C_1 \\
    r\dot{\theta}_2 C_2\\
    r\dot{\theta}_3 C_3\\
    r\dot{\theta}_4 C_4
    \end{bmatrix}=\begin{bmatrix}
    C_1 & S_1 & -w C_1 /2 &  0 \\
    C_2 & S_2 & w C_2 /2 &  0 \\
    C_3 & S_3 & (d-w/2) C_3 & - S_3 \\
    C_4 & S_4 & (d+w/2) C_4 & - S_4
    \end{bmatrix}\begin{bmatrix}
    \dot{x}_L \\
    \dot{y}_L \\
    \dot{\varphi}_L \\
    \dot{d}
    \end{bmatrix},~
% \end{equation}
% \begin{equation}
    % \label{eqn_kinematic_R}
    \begin{bmatrix}
    r\dot{\theta}_1 C_1 \\
    r\dot{\theta}_2 C_2\\
    r\dot{\theta}_3 C_3\\
    r\dot{\theta}_4 C_4
    \end{bmatrix}=\begin{bmatrix}
    C_1 & S_1 & -(d+w/2) C_1 &  S_1 \\
    C_2 & S_2 & -(d-w/2) C_2 &  S_2 \\
    C_3 & S_3 & -w C_3 /2 & 0 \\
    C_4 & S_4 & w C_4 /2 & 0
    \end{bmatrix}\begin{bmatrix}
    \dot{x}_R \\
    \dot{y}_R \\
    \dot{\varphi}_R \\
    \dot{d}
    \end{bmatrix}.
\end{equation}
\end{figure*}

\begin{figure*}[t]
    \begin{equation}
    \label{eqn_kinematic_LR}
    %\resizebox{\linewidth}{!}{$
    \begin{gathered}
    \begin{bmatrix}
        \dot{x}_L \\
        \dot{y}_L \\
        \dot{x}_R \\
        \dot{y}_R
    \end{bmatrix}
    = \dfrac{r}{2\sigma_1}
    \begin{bmatrix}
        -T_2T_3\sigma_2 + T_2T_4\sigma_3 & T_1T_3\sigma_2 - T_1T_4\sigma_3 & T_4T_1w + T_4T_2w & -T_3T_1w - T_3T_2w \\
        2(T_3d - T_4\sigma_4)& - 2( T_3\sigma_5 - T_4d) & -2 T_4w & 2T_3w \\
        -T_2T_3w - T_2T_4w & T_1T_3w + T_1T_4w & -T_4T_1\sigma_3 + T_4T_2\sigma_2 & T_3T_1\sigma_3 - T_3T_2\sigma_2 \\
        2T_2w & - 2T_1w & 2(T_1d - T_2\sigma_5) & - 2( T_1\sigma_4 - T_2d) \\
    \end{bmatrix}
    \begin{bmatrix}
        \dot{\theta}_1 \\
        \dot{\theta}_2 \\
        \dot{\theta}_3 \\
        \dot{\theta}_4
    \end{bmatrix}, 
    %$}
    \\
    \text{where } \sigma_1 = T_1T_3d - T_1T_4(d - w) - T_2T_3(d + w) + T_2T_4d,\ \sigma_2 = 2d + w,\ \sigma_3 = 2d - w,\ \sigma_4 = d - w,\ \sigma_5 = d + w.
    \end{gathered}
    \end{equation}
\end{figure*}

\begin{figure*}[t]
    \begin{equation}
    \label{eqn_kinematic}
    %\resizebox{\linewidth}{!}{$
    \begin{bmatrix}
        \dot{x}_B \\
        \dot{y}_B \\
        \dot{\varphi}_B \\
        \dot{d}
    \end{bmatrix}
    = \dfrac{r}{2\sigma_1}
    \begin{bmatrix}
        -T_2T_3\sigma_5 + T_2T_4\sigma_4 & T_1T_3\sigma_5 - T_1T_4\sigma_4 & -T_4T_1\sigma_4 + T_4T_2\sigma_5 & T_3T_1\sigma_4 - T_3T_2\sigma_5 \\
        T_2w + T_3d - T_4\sigma_4 & -T_1w - T_3\sigma_5 + T_4d & T_1d - T_2\sigma_5 - T_4w & -T_1\sigma_4 + T_2d + T_3w \\
        2T_2(T_3 - T_4) & 2T_1(-T_3 + T_4) & 2T_4(-T_1 + T_2) & 2T_3(T_1 - T_2) \\
        2(-T_2w + T_3d - T_4\sigma_4) & 2(T_1w - T_3\sigma_5 + T_4d) & 2(-T_1d + T_2\sigma_5 - T_4w) & 2(T_1\sigma_4 - T_2d + T_3w)
    \end{bmatrix}
    \begin{bmatrix}
        \dot{\theta}_1 \\
        \dot{\theta}_2 \\
        \dot{\theta}_3 \\
        \dot{\theta}_4
    \end{bmatrix}. 
    %$}
    \end{equation}    
\end{figure*}

\subsection{Control Architecture}

\begin{figure}[t]
\centering
\includegraphics[width=3.0in]{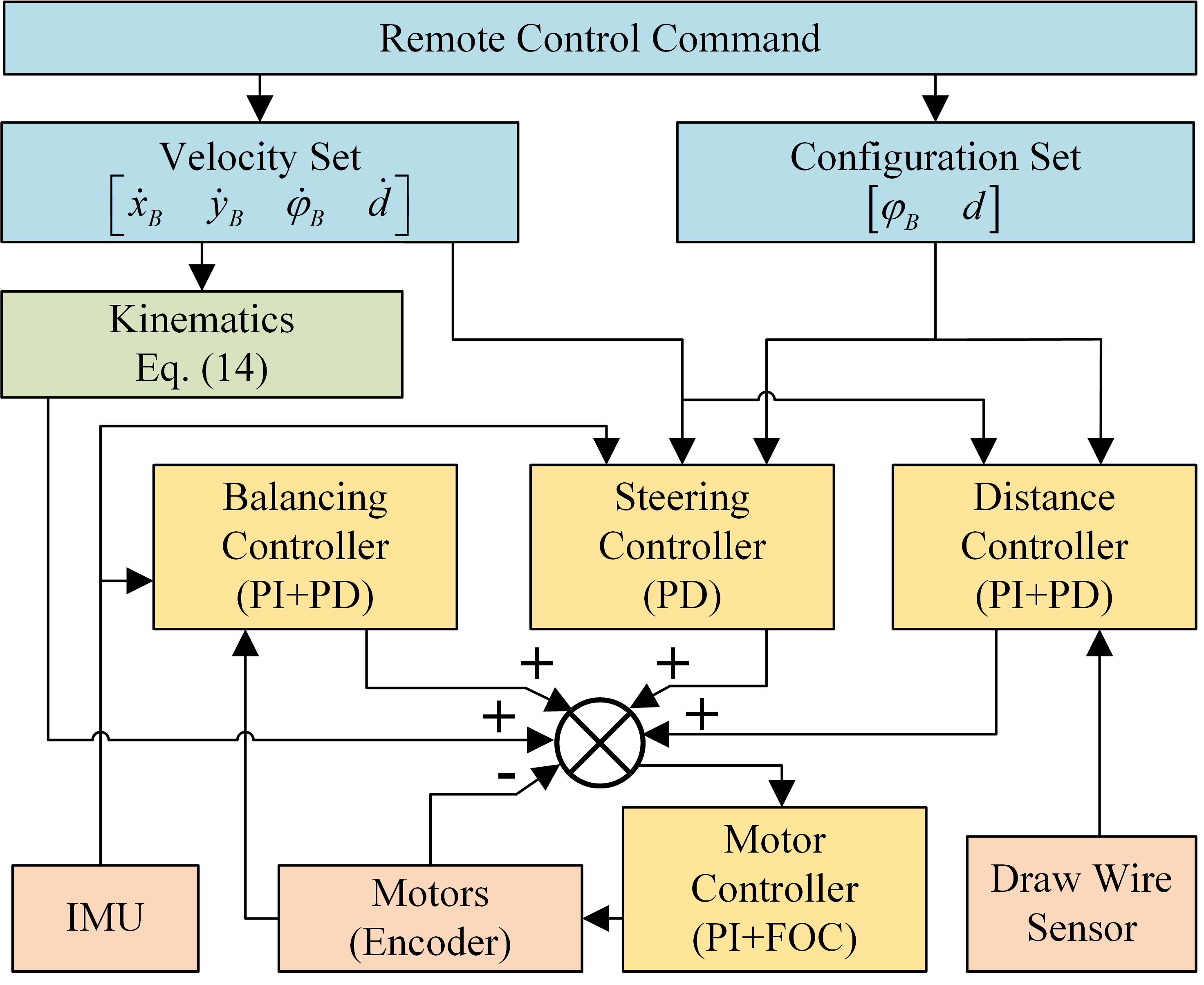}
\caption{Control architecture of the proposed prototype.}
\label{fig_control}
\end{figure}

Mecanum wheels combine driving forces in the desired direction while canceling out those in undesired directions. Fig. \ref{fig_control} illustrates the control architecture of the prototype. The robot's velocity and configuration can be adjusted with the remote control. However, the collinear arrangement of the Mecanum wheels generates additional angular velocity in certain situations, leading to significant deviations in motion control \cite{li_DesignControlTransformable_2024}. Four Proportional-Integral-Derivative (PID) controllers are implemented to maintain balance and motion control.
\subsubsection{Balancing Control} The PD controller adjusts the robot's tilt by providing corrective actions based on the pitch angle and its angular velocity from the IMU, allowing for quick responses to tilting and minimizing oscillations. The PI controller controls the robot's velocity by adjusting motor output based on velocity error. The positive feedback velocity loop enhances the system by quickly responding to any velocity changes caused by tilting. It amplifies the corrective action when the vehicle starts to tilt, helping to counteract the tilt-induced acceleration and maintain balance. Together, these components ensure the robot can rapidly adjust its angle and velocity to stay upright, even in dynamic or disturbed conditions.
\subsubsection{Steering Control} The PD controller adjusts the steering angle based on the yaw angle and its angular velocity from the IMU. The proportional term generates a corrective action proportional to the steering error, allowing quick adjustments to steer the vehicle toward the desired direction. The derivative term, on the other hand, responds to the rate of change of the steering error, helping to dampen oscillations and smooth out the steering response. This prevents the vehicle from overcorrecting and ensures that steering adjustments are accurate and stable. By balancing the contributions of the proportional and derivative terms, the PD controller enables the vehicle to steer smoothly and effectively, maintaining stability while making sharp or gradual turns.
\subsubsection{Distance Control} The PD controller adjusts the Mecanum wheels' distance based on the draw-wire sensor's length. The PI controller controls the time derivative of the draw-wire sensor values. This loop ensures that the robot's speed is moderated to avoid overshoot, with the integral term eliminating any steady-state errors. These loops enable the robot to maintain a consistent and precise distance from the target, adapting smoothly to environmental changes or disturbances while ensuring stable and accurate motion.
\subsubsection{Motor Control} The motors’ velocities are controlled by the PI controller, which receives feedback from the motor encoder. Meanwhile the DJI C620 Electronic Speed Controller (ESC) implements the current loop internally using Field-Oriented Control (FOC) to control the motor.

The speed output signals from the balancing control, steering control, and distance control, along with speed commands from the host computer or remote controller, are input into the motor control speed loop. The speed loop processes the combined speed commands and generates current commands through the PID controller. These current commands are then fed into the current loop, which controls the motor current to achieve the desired speed control.

\section{Experiments}
A series of experiments were conducted to validate the omnidirectional mobility, reconfigurability, and passability. These tests confirmed the effectiveness of the proposed prototype and the accuracy of its kinematic model, thereby validating the proposed ODD model. A demonstration video is included in the supplementary material and can be downloaded from the paper's webpage.

\subsection{Verification of Omnidirectional Mobility}

\begin{figure}[t]
    \centering
    \subfloat[]{\includegraphics[height=1.5in]{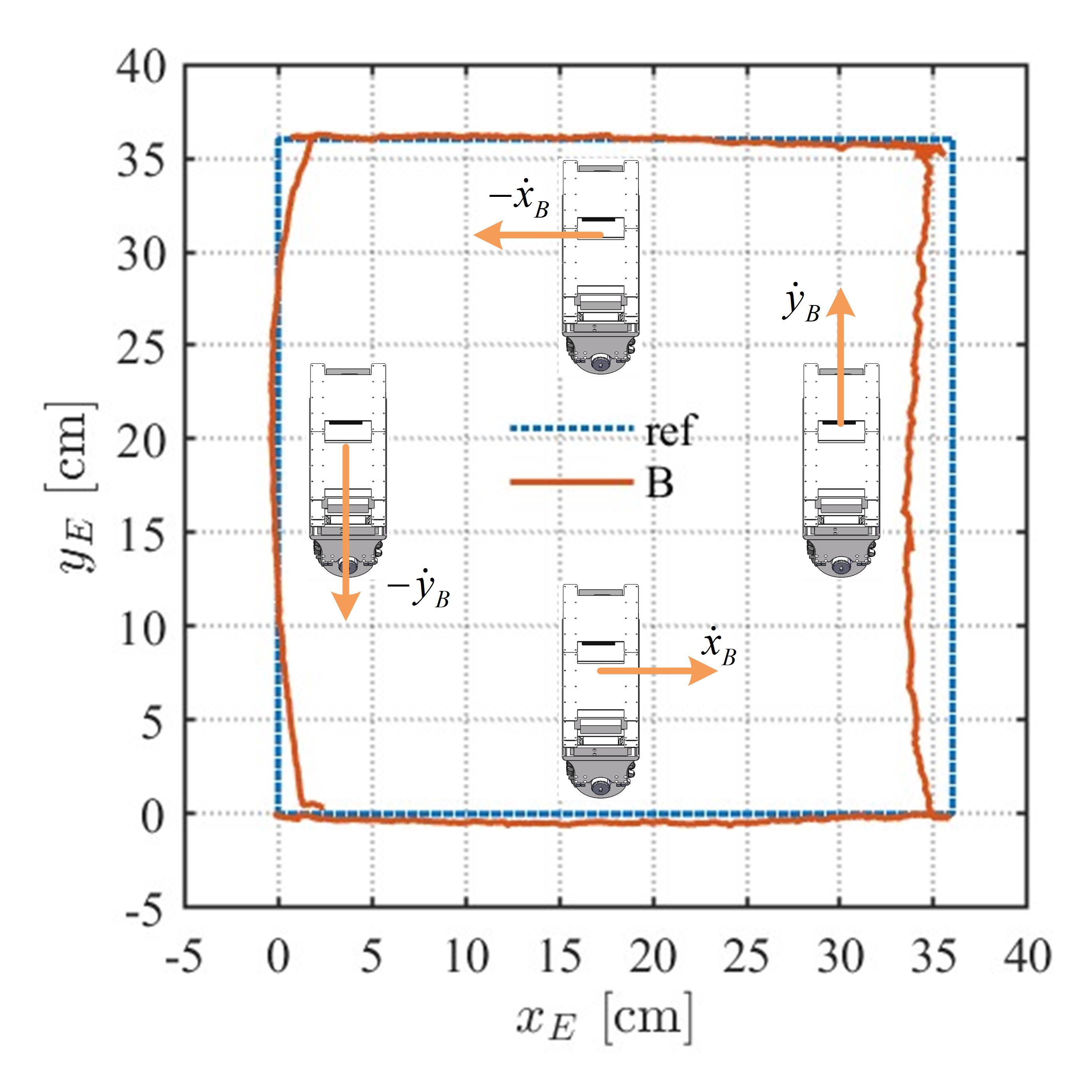}%
    \label{fig_experiment_track_XY_square}}
    \hspace{0pt}
    \subfloat[]{\includegraphics[height=1.5in]{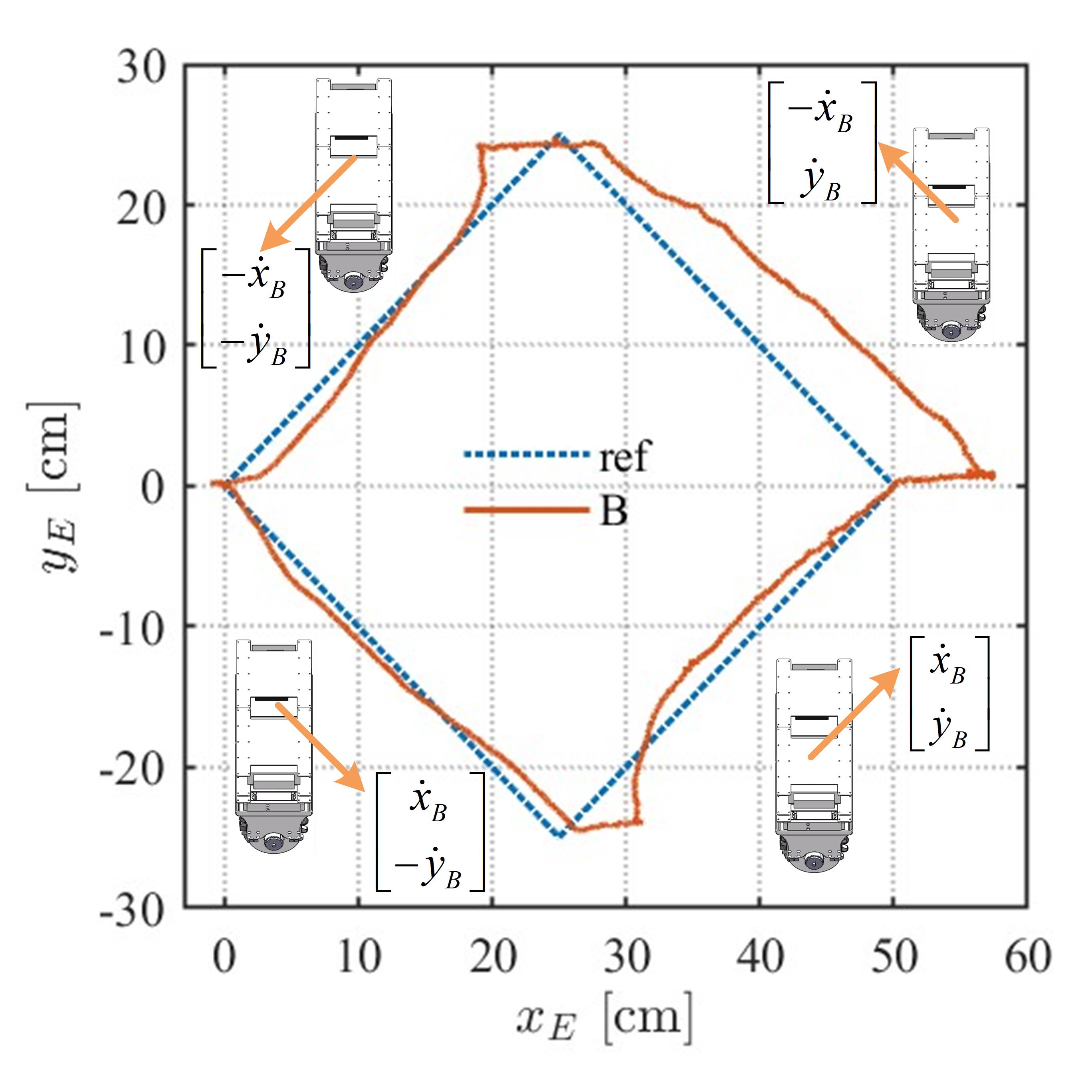}%
    \label{fig_experiment_track_XY_rhombus}}
    \hspace{0pt}
    \subfloat[]{\includegraphics[height=1.5in]{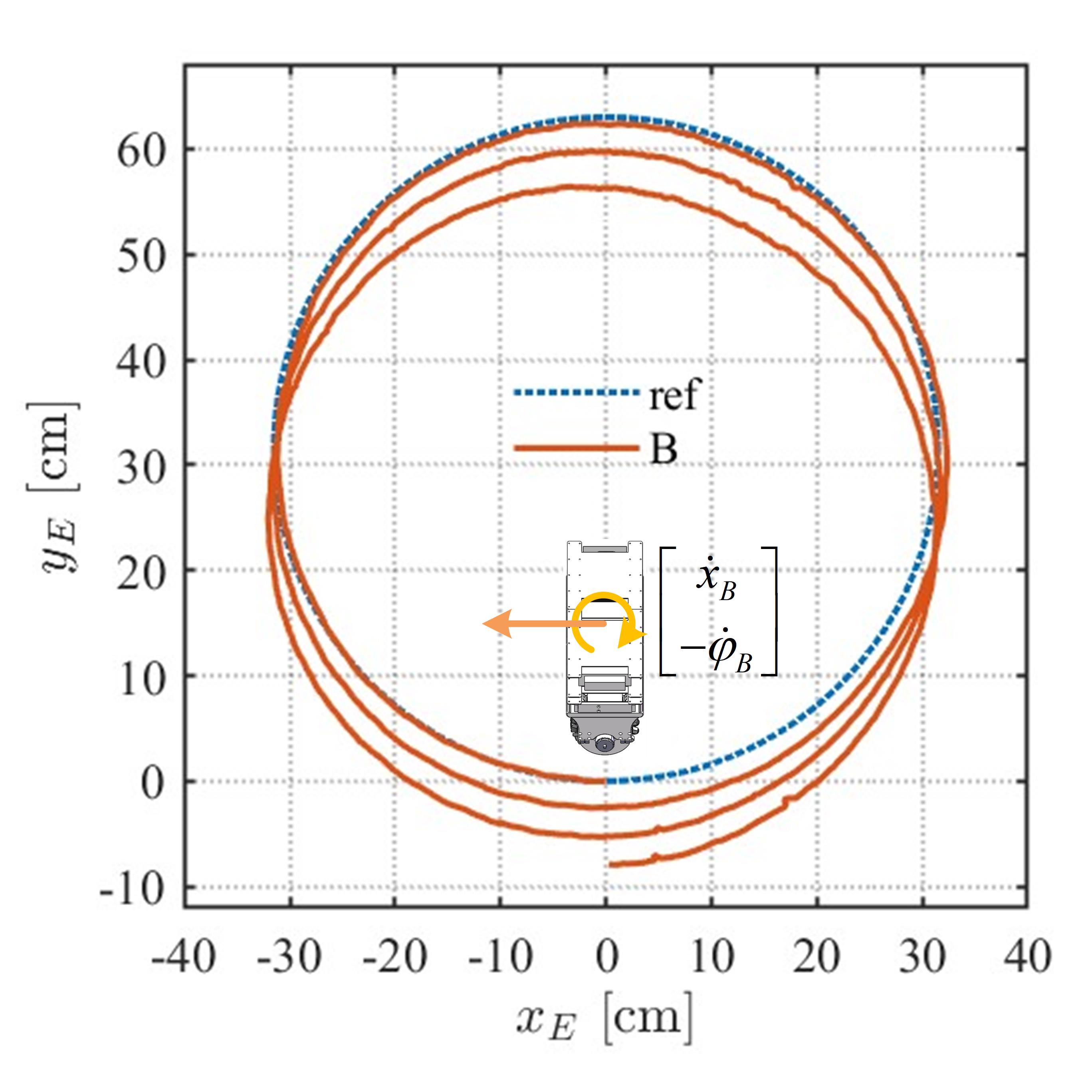}
    \label{fig_experiment_track_XZ}}
    \hspace{0pt}
    \subfloat[]{\includegraphics[height=1.5in]{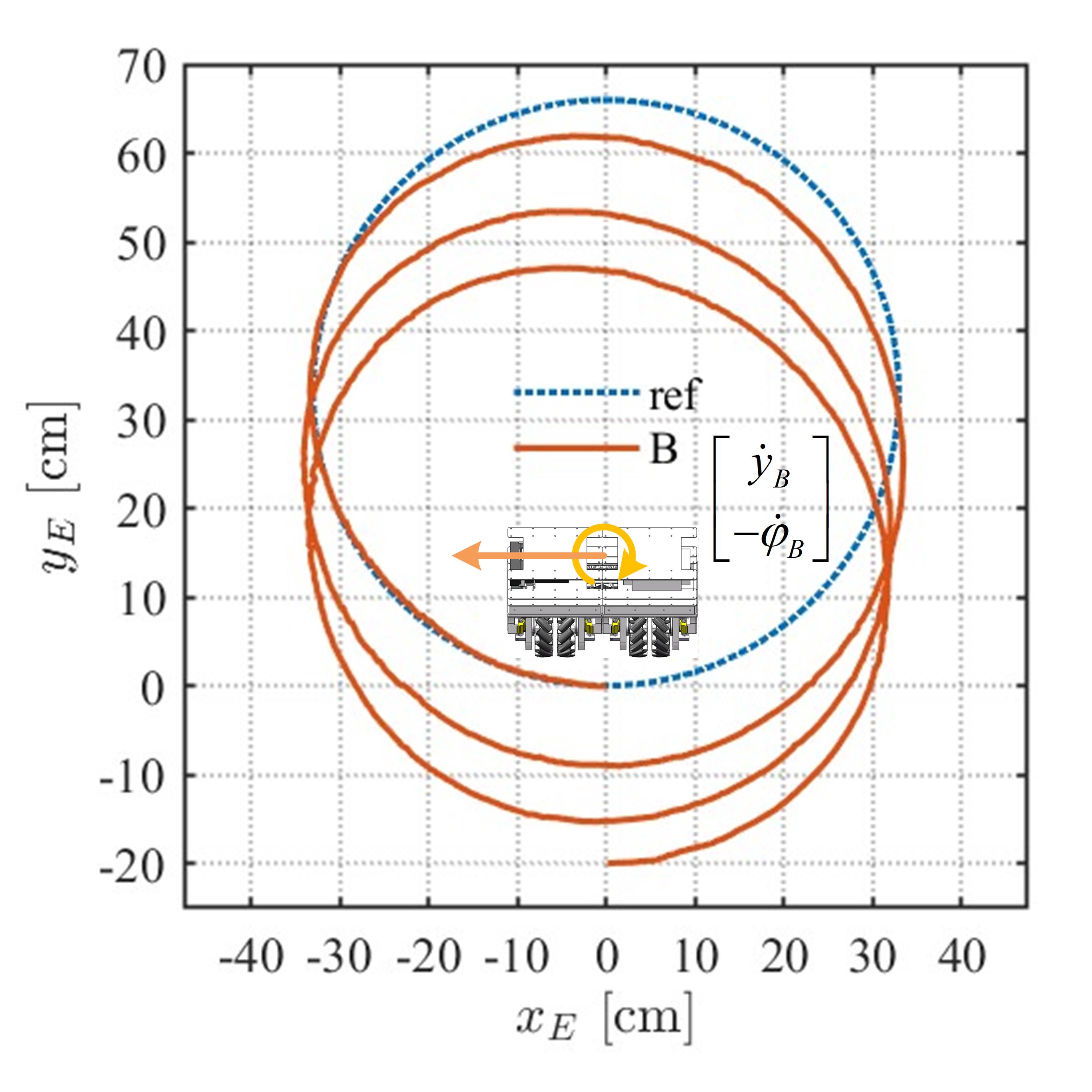}
    \label{fig_experiment_track_YZ}}
    \caption{Paths of center of the prototype. (a) Square: 
Either $\dot{x}_B$ or $\dot{y}_B$. (b) Rhombus: Both $\dot{x}_B$ and $\dot{y}_B$. (C) Circle: Both $\dot{x}_B$ and $\dot{\varphi}_B$. (d) Circle: Both $\dot{y}_B$ and $\dot{\varphi}_B$. The starting coordinates are (0,0).}
    \label{fig_experiment_track_XYZ}
\end{figure}

Omnidirectional movement involves three degrees of freedom: translation along the $B_x$ and $B_y$ axes, and rotation around the $B_z$ axis. Four experiments were conducted to verify the accuracy of the prototype's movement according to commands without position compensation. Fig. \ref{fig_experiment_track_XYZ} shows the motion trajectories of the prototype's center, with data collected by the OptiTrack motion capture system.

In Fig. \ref{fig_experiment_track_XYZ}\subref{fig_experiment_track_XY_square}, velocities $\dot{x}_B$ and $\dot{y}_B$ along the $B_x$ and $B_y$ axes were applied for a given duration $t$. Theoretically, this should produce a square trajectory. The curvature is mainly due to slight tilting caused by self-balancing along the $B_x$ axis during $B_y$ movement.

Fig. \ref{fig_experiment_track_XYZ}\subref{fig_experiment_track_XY_rhombus} shows the result when velocities and durations are applied simultaneously to the $B_x$ and $B_y$ axes, theoretically producing a rhombus trajectory. The curvature is mainly due to slight tilting caused by self-balancing along the $B_x$ axis during $B_y$ movement.

Fig. \ref{fig_experiment_track_XYZ}\subref{fig_experiment_track_XZ} and Fig. \ref{fig_experiment_track_XYZ}\subref{fig_experiment_track_YZ} shows the result when velocities along the $B_x$ axis or $B_y$ axis, and angular velocities around the $B_z$ axis are applied, theoretically producing a circular trajectory. The deviation between the starting and ending coordinates in Fig. \ref{fig_experiment_track_XYZ}\subref{fig_experiment_track_YZ} is larger than that observed in Fig. \ref{fig_experiment_track_XYZ}\subref{fig_experiment_track_XZ}. 

\begin{figure}[t]
    \centering
    \subfloat[]{\includegraphics[height=1.5in]{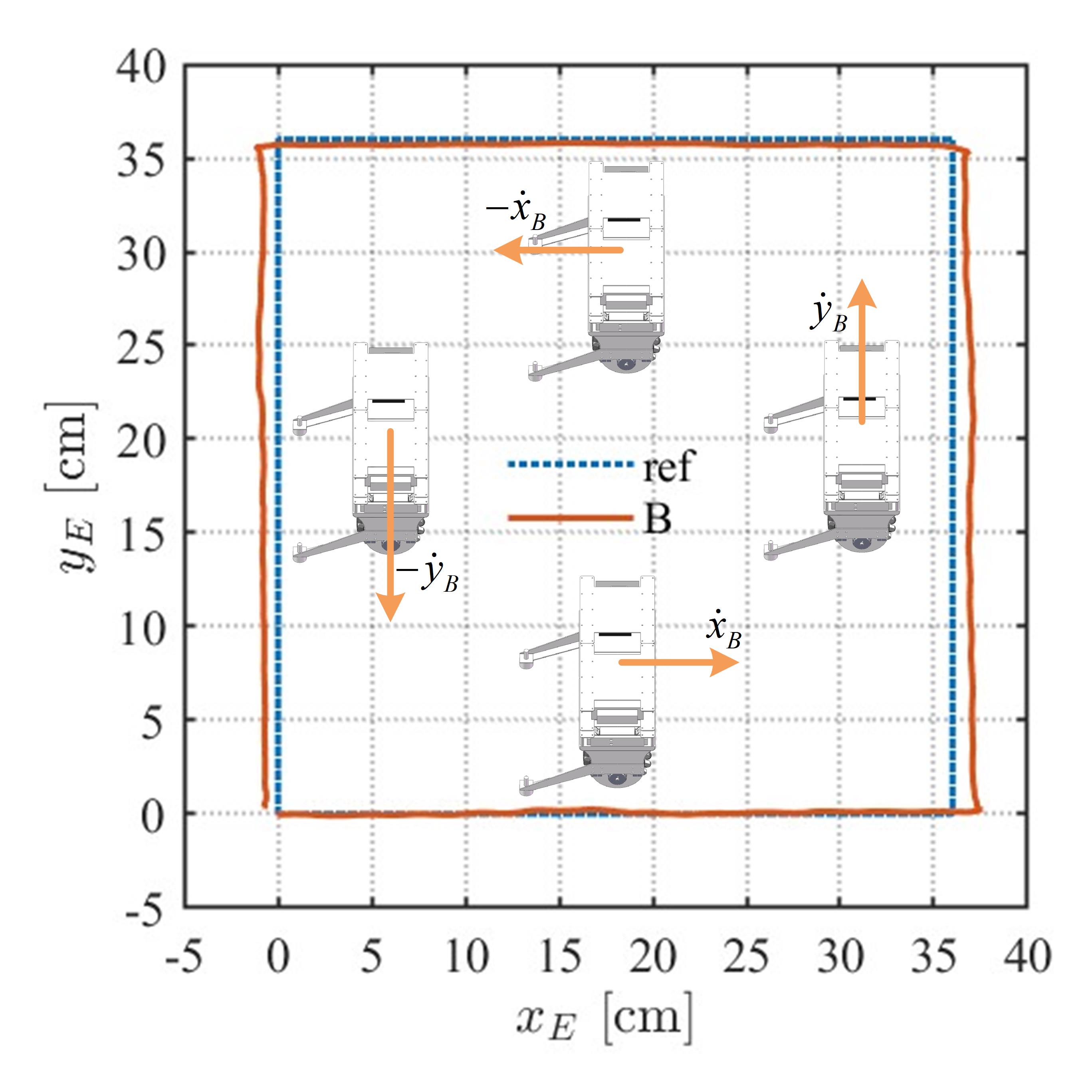}%
    \label{fig_experiment_track_XY_square2}}
    \hspace{0pt}
    \subfloat[]{\includegraphics[height=1.5in]{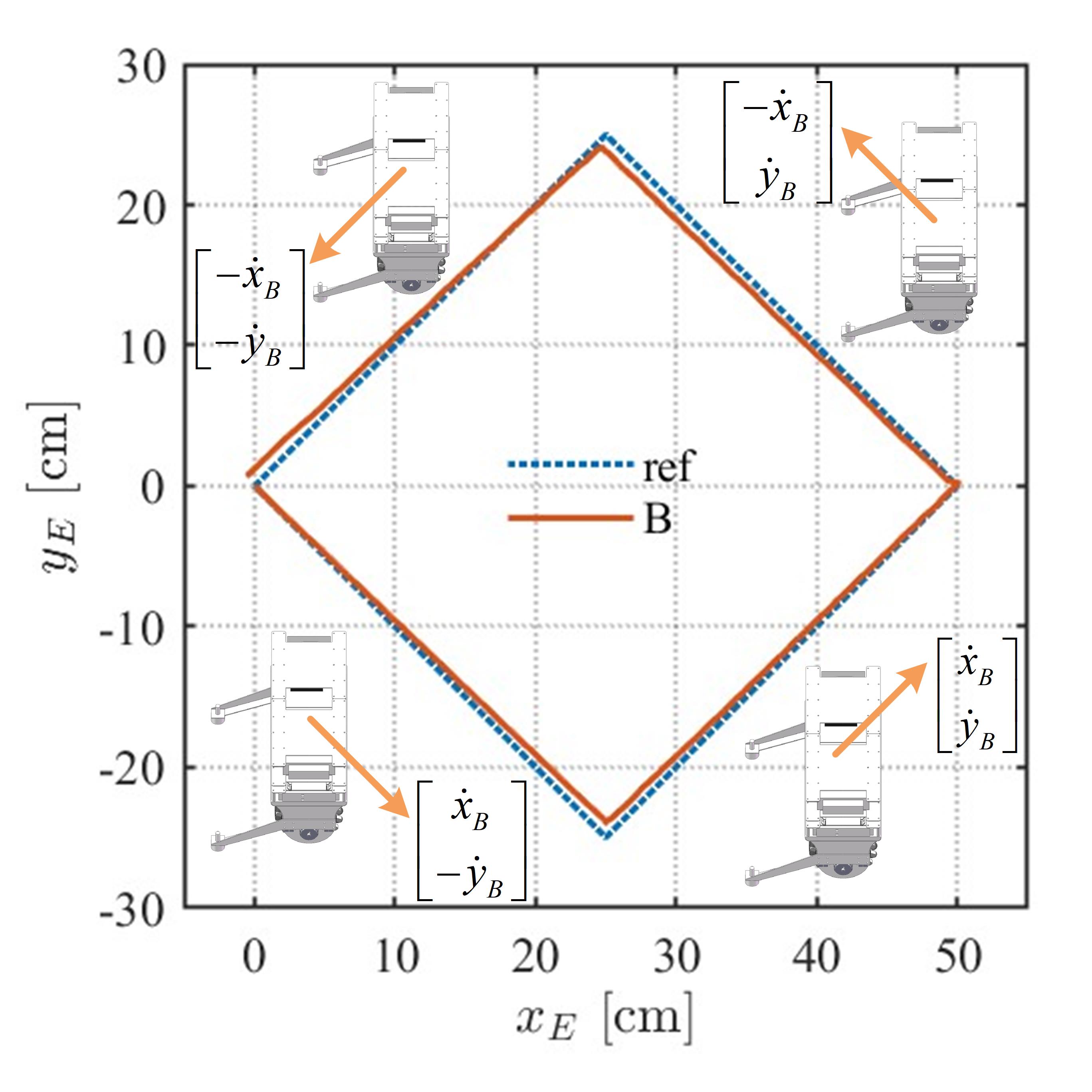}%
    \label{fig_experiment_track_XY_rhombus2}}
    \hspace{0pt}
    \subfloat[]{\includegraphics[height=1.5in]{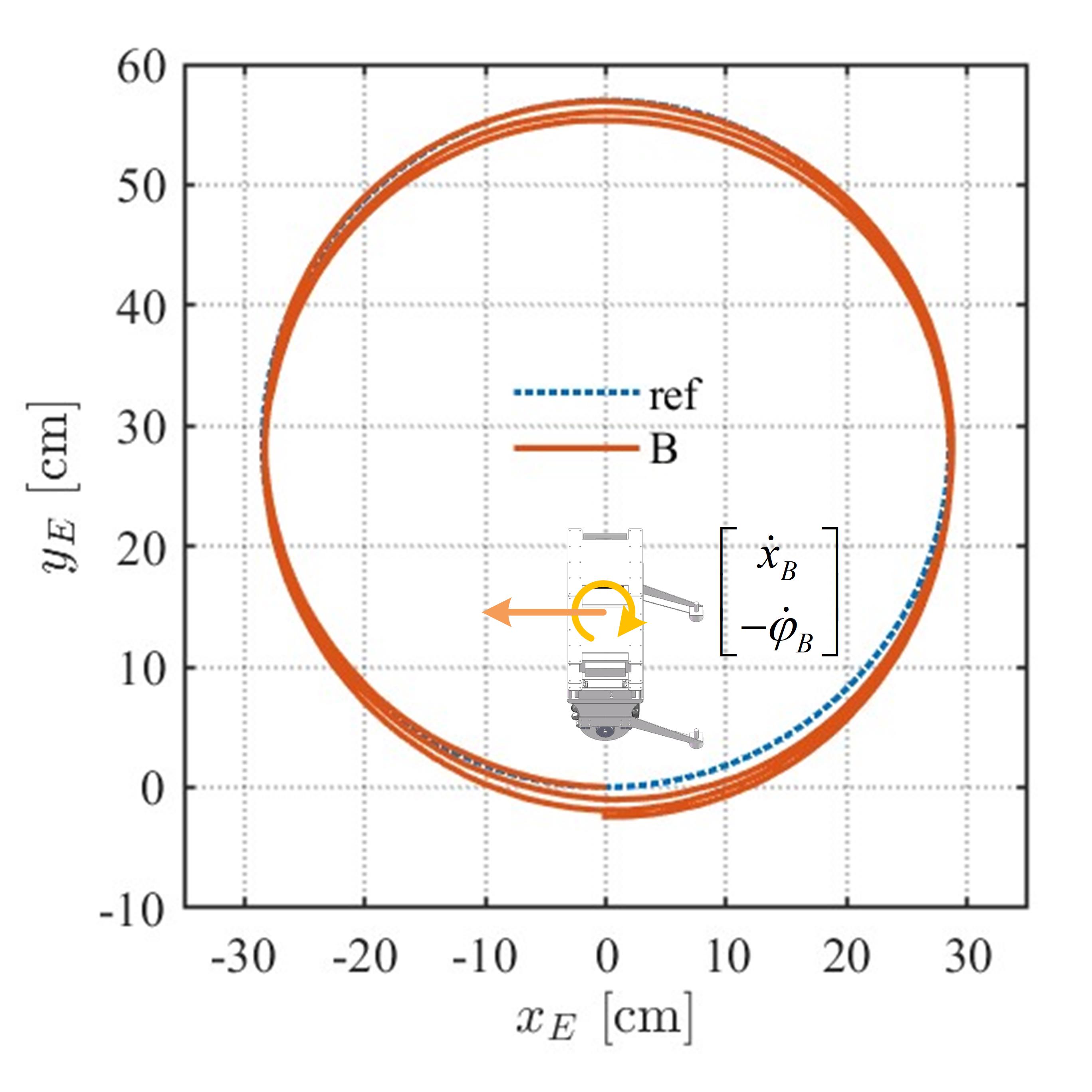}
    \label{fig_experiment_track_XZ2}}
    \hspace{0pt}
    \subfloat[]{\includegraphics[height=1.5in]{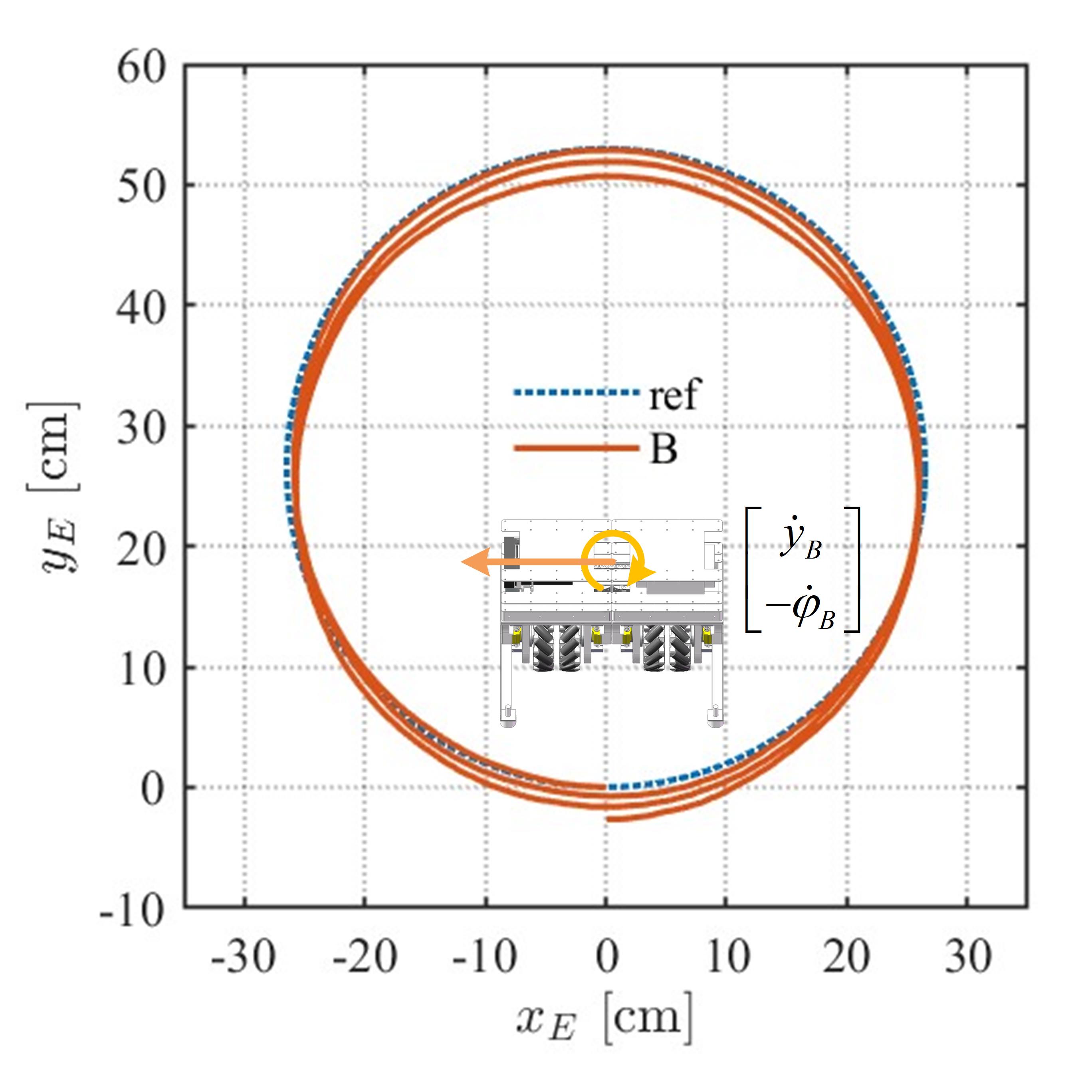}
    \label{fig_experiment_track_YZ2}}
    \caption{Paths of center of the prototype with support ball casters. (a) Square: 
Either $\dot{x}_B$ or $\dot{y}_B$. (b) Rhombus: Both $\dot{x}_B$ and $\dot{y}_B$. (C) Circle: Both $\dot{x}_B$ and $\dot{\varphi}_B$. (d) Circle: Both $\dot{y}_B$ and $\dot{\varphi}_B$. The starting coordinates are (0,0).}
    \label{fig_experiment_track_XYZ2}
\end{figure}

Due to the significant errors introduced by self-balancing, eliminating these errors helps in analyzing other error factors. Thus, two support ball casters were added to the prototype to maintain stability in the $B_x$ direction without self-balancing. The same trajectory experiment was conducted as shown in Fig. \ref{fig_experiment_track_XYZ}, with results presented in Fig. \ref{fig_experiment_track_XYZ2}.

As shown in Fig. \ref{fig_experiment_track_XYZ2}\subref{fig_experiment_track_XY_square2}, after eliminating the influence of self-balancing, the motion in the $B_x$ direction remains more stable than in the $B_y$ direction. A preliminary analysis suggests that when moving along the $B_x$ direction, frictional forces generated by the four Mecanum wheels in the $B_y$ direction are coaxial and can cancel each other out. Even if dynamic disturbances cause a non-zero resultant force in the $B_y$ direction, it does not lead to deviations in the angle $\varphi_B$. However, when moving along the $B_y$ direction, the forces generated by the four Mecanum wheels in the $B_x$ direction are parallel but not coaxial, resulting in torque that disturbs the angle $\varphi_B$, making the trajectory deviation more noticeable.Fig. \ref{fig_experiment_track_XYZ2}\subref{fig_experiment_track_XY_rhombus2} shows that errors in the $B_x$ direction at transitions of the straight-line trajectory, caused by inertia, are eliminated. However, as in Fig. \ref{fig_experiment_track_XYZ2}\subref{fig_experiment_track_XY_square2}, motion inconsistency between the $B_x$ and $B_y$ directions persists for the same reasons.

Compared to Fig. \ref{fig_experiment_track_XYZ}\subref{fig_experiment_track_XZ} and Fig. \ref{fig_experiment_track_XYZ}\subref{fig_experiment_track_YZ}, trajectory deviations in Fig. \ref{fig_experiment_track_XYZ2}\subref{fig_experiment_track_XZ} and Fig. \ref{fig_experiment_track_XYZ2}\subref{fig_experiment_track_YZ} are significantly smaller, with end-point deviations of 2.42 cm and 2.67 cm, and one-loop deviations of 0.99 cm and 0.74 cm, respectively. The difference in displacement between the two is minimal, suggesting that the regular deviations may be caused by gravitational components due to ground inclination. The larger deviation in Fig. \ref{fig_experiment_track_XYZ}\subref{fig_experiment_track_YZ} compared to Fig. \ref{fig_experiment_track_XYZ}\subref{fig_experiment_track_XZ} is due to the same reason as in the Square and Rhombus trajectory experiments: motion along the $B_y$ direction is affected by additional torque.

In Fig. \ref{fig_experiment_track_XYZ}, the prototype with self-balancing control is dynamically stable but shows larger deviations under disturbances like additional torque and ground inclination, compared to Fig. \ref{fig_experiment_track_XYZ2}.

\subsection{Verification of Reconfigurability in Motion}

\begin{figure}
    \centering
    \subfloat[]{\includegraphics[height=1.5in]{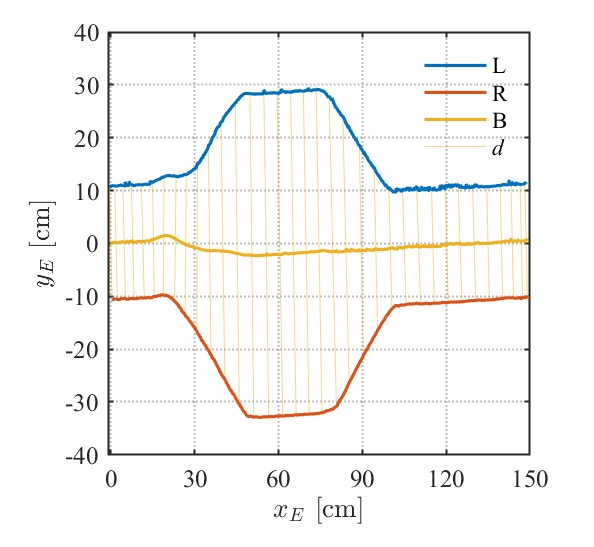}%
    \label{fig_experiment_track_XD_XY}}
    \hspace{0pt}
    \subfloat[]{\includegraphics[height=1.5in]{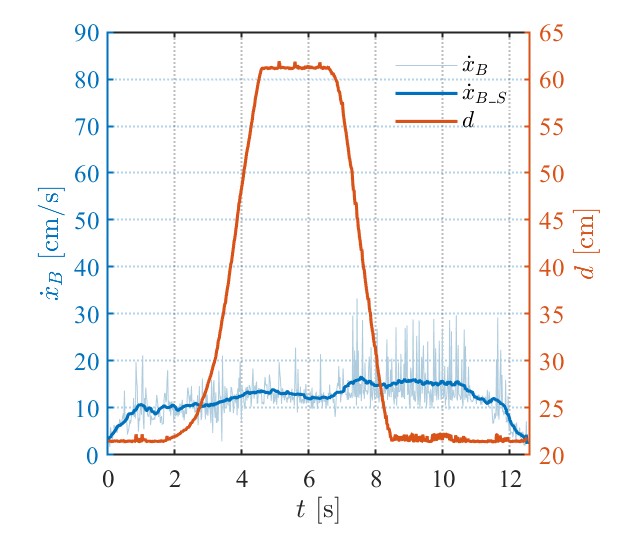}%
    \label{fig_experiment_track_XD_d}}
    \hspace{0pt}
    \subfloat[]{\includegraphics[height=1.5in]{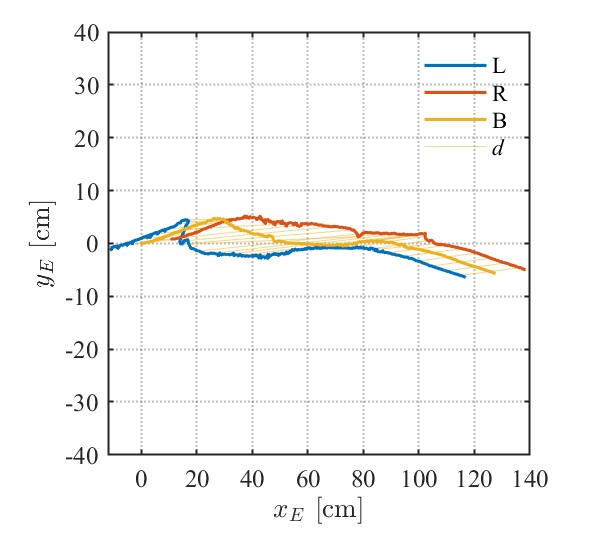}
    \label{fig_experiment_track_YD_XY}}
    \hspace{0pt}
    \subfloat[]{\includegraphics[height=1.5in]{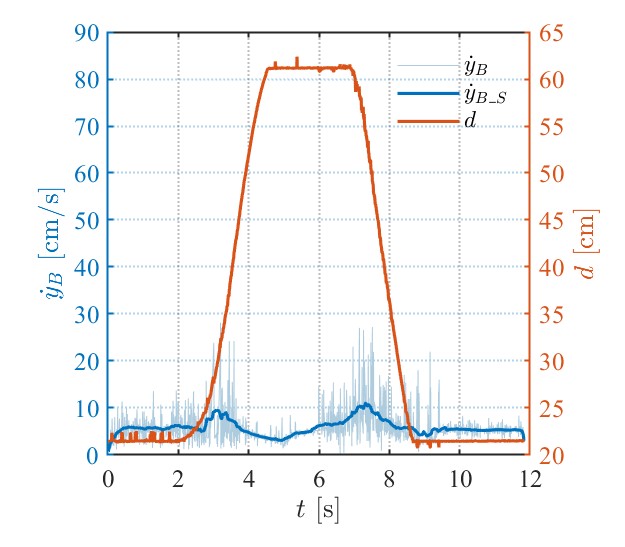}
    \label{fig_experiment_track_YD_d}}
    \hspace{0pt}
    \subfloat[]{\includegraphics[height=1.5in]{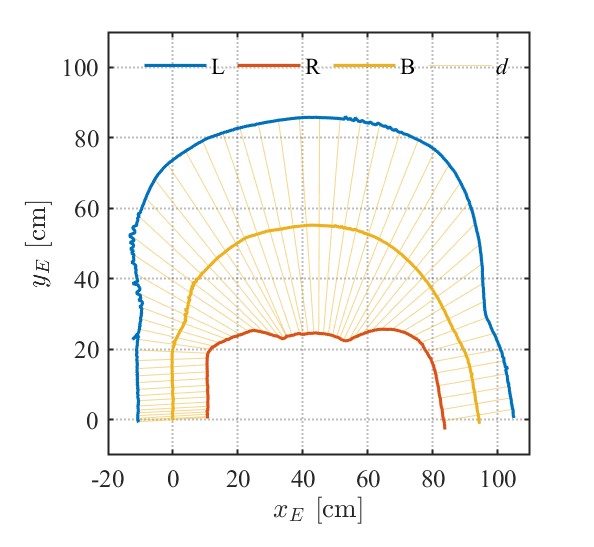}
    \label{fig_experiment_track_ZD_XY}}
    \hspace{0pt}
    \subfloat[]{\includegraphics[height=1.5in]{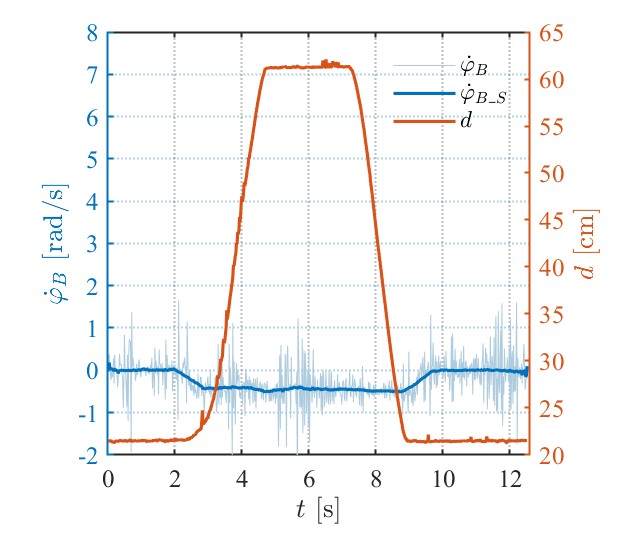}
    \label{fig_experiment_track_ZD_d}}
    \caption{Effect of $d$ variation on trajectory and speed. (a) and (b) are of mobile along $B_x$. (c) and (d) are of mobile along $B_y$. (e) and (f) are of mobile along $B_x$ and around $B_z$. $\dot{x}_{B\_S}$,$\dot{y}_{B\_S}$,$\dot{\varphi}_{B\_S}$ represent smoothed data. The starting coordinates are (0,0).}
   \label{fig_experiment_track_D}
\end{figure}

To validate the impact of varying wheel spacing $d$ on movement, three experiments were designed. These experiments involved changing the wheel spacing $d$ while moving straight along the $B_x$ and $B_y$ axes and rotating around the $B_z$ axis. An OptiTrack motion capture system recorded the positions of the left wheel group center $L$, the right wheel group center $R$, and the prototype center $B$. Using the positional data, the wheel spacing $d$ and the movement speed of the prototype center were calculated, as shown in Fig. \ref{fig_experiment_track_D}.

When moving along the $B_x$ axis and changing the wheel spacing $d$, the trajectories of the prototype center and the left and right wheel groups are shown in Fig. \ref{fig_experiment_track_D}\subref{fig_experiment_track_XD_XY}. The trajectory of the prototype center essentially follows a straight line along the X-axis. However, when the wheel spacing increases, the trajectory deviates. The primary reasons for this deviation are the difference in mass between the left and right sides, which leads to variations in friction and inertia, as well as the additional torque generated by changes in $d$. Fig. \ref{fig_experiment_track_D}\subref{fig_experiment_track_XD_d} shows that $\dot{x}_B$ is minimally affected by changes in $d$, The primary error source is self-balancing in the $B_x$ direction.

When moving along the $B_y$ axis and changing the wheel spacing $d$, the trajectory of the prototype center is shown in Fig. \ref{fig_experiment_track_D}\subref{fig_experiment_track_YD_XY}. The trajectory of the prototype center exhibits significant errors along the Y-axis. The main reason is that dynamic balancing in the $B_x$ direction induces speed and displacement in the $B_x$ direction, particularly when $d$ changes, affecting the dynamic balance. In Fig. \ref{fig_experiment_track_D}\subref{fig_experiment_track_YD_d}, it is evident that the speed $\dot{y}_B$ is significantly influenced by changes in $d$. It is because the preset $\dot{d}/2$ is greater than the preset $\dot{y}_B$. As $d$ increases, the left wheel group needs to change its direction of motion. When the speed of the left wheel group decreases to zero, significant static friction occurs, which is greater than the dynamic friction of the right wheel group. This results in $\dot{y}_B$ equaling the preset $\dot{d}/2$, which is greater than the preset $\dot{y}_B$. Conversely, when $d$ decreases, the right wheel group needs to change its direction of motion, encountering a similar situation as the left wheel group.

When movement along the $B_x$ axis and rotation around the $B_z$ axis occur simultaneously, changing the wheel spacing $d$, the trajectory of the prototype center is shown in Fig. \ref{fig_experiment_track_D}\subref{fig_experiment_track_ZD_XY}. Initially, only $\dot{x}_B$ is present, then $\dot{\varphi}_B$ is added, and finally, $\dot{\varphi}_B$ is removed, leaving only $\dot{x}_B$. The trajectory of the prototype center is minimally affected by changes in $d$. Fig. \ref{fig_experiment_track_D}\subref{fig_experiment_track_ZD_d} demonstrates the predetermined changes in $\dot{\varphi}_B$, which are largely unaffected by changes in $d$. However, the dynamics analysis shows that the simultaneous presence of $\dot{\varphi}_B$ and $\dot{d}$ induces Coriolis acceleration, disturbing the system's motion. Additionally, changes in $d$ affect the moment of inertia $I$, which in turn influences the torque required for rotation.

\subsection{Demonstration of the Passability}

\begin{figure}[t]
\centering
\includegraphics[width=3.3in]{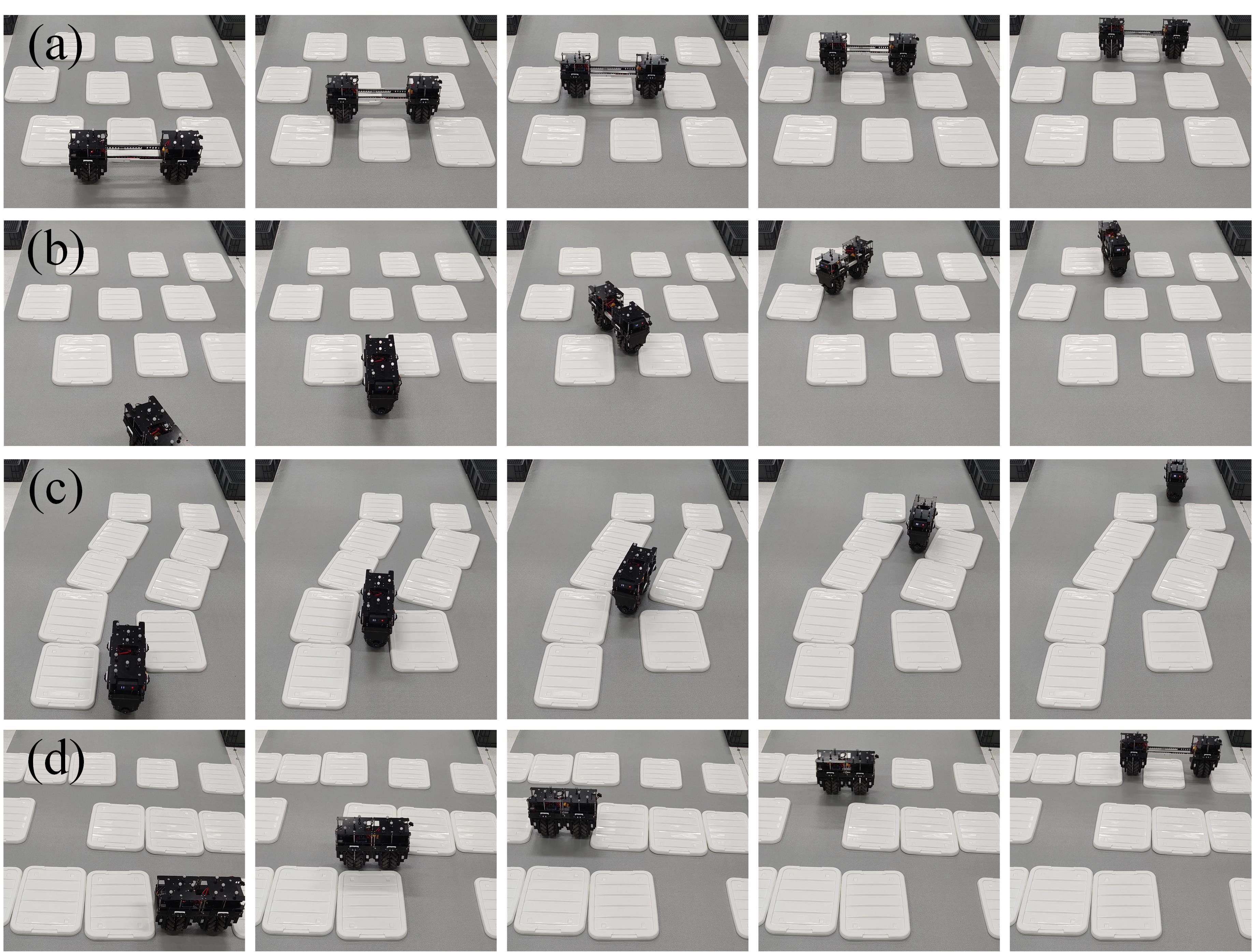}
\caption{Demonstration of the passability. (a) Fixed orientation. (b) Lateral movement. (c) Narrow passage. (d) Changing $d$ in the path.}
\label{fig_demo}
\end{figure}

To demonstrate the prototype's traversability, the following experiments were conducted. Fig. \ref{fig_demo} shows images extracted from the demonstration video. Fig. \ref{fig_demo}(a) illustrates the prototype navigating through a complex obstacle environment with an appropriate wheel distance, maintaining a constant orientation. Fig. \ref{fig_demo}(b) shows the prototype navigating the same obstacles as in Fig. \ref{fig_demo}(a) by combining omnidirectional movement and varying wheel distance. Due to self-balancing control in the $B_x$ direction, the prototype oscillates to maintain balance, making control more challenging.Fig. \ref{fig_demo}(c) shows the prototype using omnidirectional movement to pass through a narrow passage. Fig. \ref{fig_demo}(d) shows the prototype adjusting wheel distance during obstacle traversal to adapt to different channels.
 
% \begin{figure}[!t]
%     \centering
%     \subfloat[]{\includegraphics[width=3.3in]{fig/demo_1_ss.jpg}%
%     \label{fig_demo_1}}
%     \hspace{0pt}
%     \subfloat[]{\includegraphics[width=3.3in]{fig/demo_2_ss.jpg}%
%     \label{fig_demo_2}}
%     \hspace{0pt}
%     \subfloat[]{\includegraphics[width=3.3in]{fig/demo_3_ss.jpg}%
%     \label{fig_demo_3}}
%     \hspace{0pt}
%     \subfloat[]{\includegraphics[width=3.3in]{fig/demo_4_ss.jpg}%
%     \label{fig_demo_4}}    
%     \caption{Demonstration of the passability.(a) Fixed orientation. (b) Lateral movement. (c) Narrow passage. (d) Changing $d$ in the path.}
%     \label{fig_demo}
% \end{figure}

% \begin{figure}[!t]
%     \centering
%     \subfloat[]{\includegraphics[width=3.3in]{fig/demo_1_sss.jpg}%
%     \label{fig_demo_1}}
%     \hspace{0pt}
%     \subfloat[]{\includegraphics[width=3.3in]{fig/demo_2_sss.jpg}%
%     \label{fig_demo_2}}
%     \hspace{0pt}s
%     \subfloat[]{\includegraphics[width=3.3in]{fig/demo_3_sss.jpg}%
%     \label{fig_dsemo_3}}
%     \hspace{0pt}
%     \subfloat[]{\includegraphics[width=3.3in]{fig/demo_4_sss.jpg}%
%     \label{fig_demo_4}}    
%     \caption{Demonstration of the passability.(a) Fixed orientation. (b)s Lateral movement. (c) Narrow passage. (d) Changing $d$ in the path.}
%     \label{fig_demo}
% \end{figure}

\section{Conclusion}
In this study, we developed a novel Omni Differential Drive (ODD) wheeled mobility inspired by human movements and designed a prototype based on collinear Mecanum wheels to implement and verify the ODD. The ODD enables simultaneous reconfiguration and omnidirectional mobility for wheeled robots, meeting requirements for passability, agility, and stability in human living environments. Moreover, The kinematics of the ODD were modeled and used to establish the prototype's kinematic model. A Parallel Cascade PID control system was designed for the prototype, and the models of both the ODD and prototype were experimentally validated.

Future work will apply the ODD to more chassis configurations, given its unique characteristics. For instance, adding castors can transform it into a reconfigurable, omnidirectional multi-wheeled vehicle without requiring dynamic balancing. It can also be applied to wheel-legged robots, where omnidirectional movement and adjustable wheel spacing work with leg joints to perform more complex actions. Additional, we will design lighter, more compact active omnidirectional wheels driven by the ODD model. Path planning for mobile robots using the ODD model will be another focus of future research. A more detailed dynamic analysis of the reconfigurable collinear Mecanum wheels will be conducted to identify the sources of errors. Based on this analysis, more advanced dynamic control strategies will be developed to address these issues.

% \balance
\bibliographystyle{IEEEtran}
\bibliography{IEEEabrv, bibliography}
% \input{main.bbl}

% \vfill
\end{document}